\documentclass[letterpaper, 10 pt, conference]{ieeeconf}  %% Comment this line out if you need a4paper
\IEEEoverridecommandlockouts                 %% This command is only needed if you want to use the \thanks command
\usepackage{epsfig} % for postscript graphics files
\usepackage{times} % assumes new font selection scheme installed
\usepackage{amsmath} % assumes amsmath package installed
\usepackage{amssymb}  % assumes amsmath package installed
\usepackage{algorithm}
\usepackage{algorithmic}
\usepackage{url}

\title{\LARGE \bf Real-Time 6D Object Pose Estimation on CPU}

\author{Yoshinori Konishi$^{1}$, Kosuke Hattori$^{1}$ and Manabu Hashimoto$^{2}$
\thanks{$^{1}$Y. Konishi and K. Hattori are with OMRON Corporation, Kyoto, Japan.  Konishi's current affiliation is SenseTime Japan Ltd., Kyoto, Japan {\tt\small konishi@sensetime.jp}.}
\thanks{$^{2}$M. Hashimoto is with Dept. Engineering, Chukyo Univ., Nagoya, Japan.}}

\begin{document}

\maketitle
\thispagestyle{empty}
\pagestyle{empty}

%%%%%%%%%%%%%%%%%%%%%%%%%%%%%%%%%%%%%%%%%%%%%%%%%%%%%%%%%%%%%%%%%%%%%%%%%%%%%%%%
\begin{abstract}
We propose a fast and accurate 6D object pose estimation from a RGB-D image.  Our proposed method is template matching based and consists of three main technical components, PCOF-MOD (multimodal PCOF), balanced pose tree (BPT) and optimum memory rearrangement for a coarse-to-fine search.  Our model templates on densely sampled viewpoints and PCOF-MOD which explicitly handles a certain range of 3D object pose improve the robustness against background clutters.  BPT which is an efficient tree-based data structures for a large number of templates and template matching on rearranged feature maps where nearby features are linearly aligned accelerate the pose estimation.  The experimental evaluation on tabletop and bin-picking dataset showed that our method achieved higher accuracy and faster speed in comparison with state-of-the-art techniques including recent CNN based approaches.  Moreover, our model templates can be trained solely from 3D CAD in a few minutes and the pose estimation run in near real-time (23 fps) on CPU.  These features are suitable for any real applications.
\end{abstract}
%%%%%%%%%%%%%%%%%%%%%%%%%%%%%%%%%%%%%%%%%%%%%%%%%%%%%%%%%%%%%%%%%%%%%%%%%%%%%%%%
\section{Introduction}
Detecting 3D position and pose (6 degrees of freedom) of object instances is one of the essential techniques in computer vision and is widely used in various applications such as robotic manipulation and augmented reality.  These real applications require handling of various objects including texture-less and simple-shaped objects, fast processing time on poor computer resources, and robustness against background clutters and partial occlusions.  Additionally, immediate on-site training is also required because the target objects are different in every application.

In recent years, many 6D object pose estimation algorithms based on CNN have been proposed \cite{Kehl2016, Kehl2017, Rad2017, Sundermeyer2018, Tekin2018}.  Though they are fast and robust, they require high-performance but costly GPU and many real/synthetic training samples.  It is not realistic for each user to collect and annotate many training samples and run training programs for long hours in every application.

The template matching based approaches \cite{Hinterstoisser2012b, Kehl2015} and local descriptor based approaches \cite{Drost2010, Hinterstoisser2016} which use only geometric features can be trained from 3D CAD of objects.  Though they do not require any additional training samples and the training is done in a shorter time, their estimation accuracy and robustness are often inferior to those of CNN based approaches.

We propose template matching based 6D object pose estimation algorithm from a RGB-D image.  Though recent CNN based approaches recover 6D object pose only from a RGB image, the pose refinement using depth information is essential for estimation of precise 6D pose especially for robotic applications.  Our proposed algorithm can estimate precise 6D pose (pose errors are less than 1 mm) in real-time on CPU.  Moreover, it is robust enough for handling a bin-picking scene and the model templates are trained solely from 3D CAD of a target object in a few minutes.

\begin{figure}[t]
 \begin{center}
  \begin{tabular}{c}
   \begin{minipage}{4.1cm}
    \begin{center}
     \includegraphics[clip, width=4.1cm]{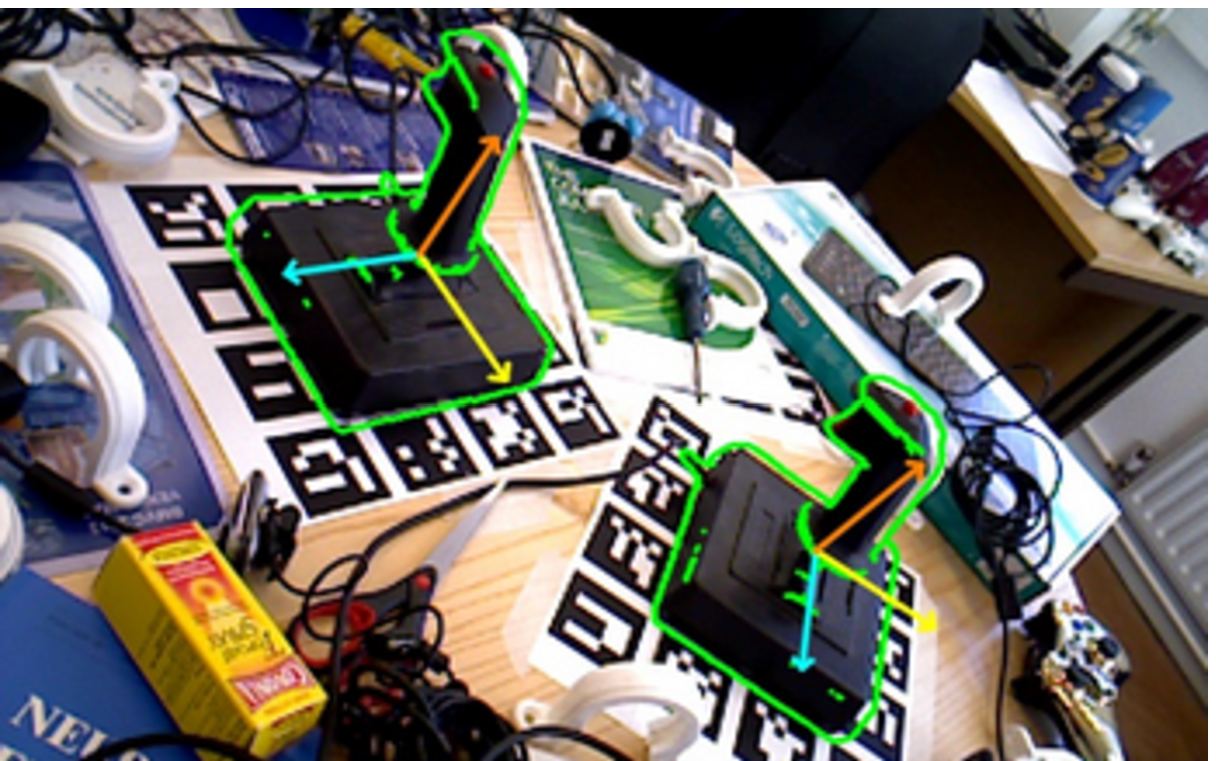}
    \end{center}
   \end{minipage}
   \begin{minipage}{4.3cm}
    \begin{center}
     \includegraphics[clip, width=4.3cm]{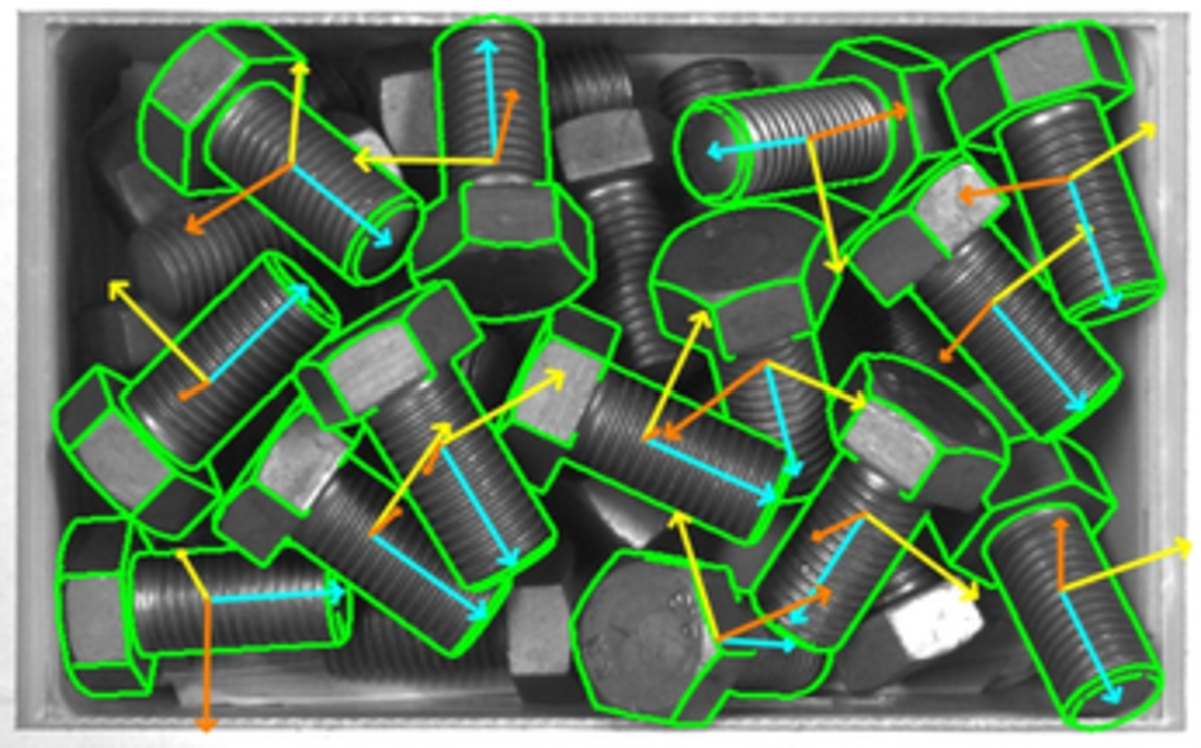}
    \end{center}
   \end{minipage}
  \end{tabular}
 \end{center}
 \caption{The estimated 6D object pose by our proposed method on a tabletop (left) and a bin-picking (right) scene.  It takes approximately 43 ms and 90 ms on CPU for each scene.  Moreover, the model training requires only 3D CAD of the object and it takes just a few minutes.}
 \label{fig:headline}
\end{figure}

Our proposed algorithm consists of three main technical components and the contributions are as follows: 
\begin{itemize}
 \item[\textbullet] First component is PCOF-MOD (multi-MODal Perspectively Cumulated Orientation Feature).  PCOF-MOD relaxes the matching conditions only for small changes in 3D object pose without increasing false positives under cluttered background.
 \item[\textbullet] Second component is BPT (Balanced Pose Tree).  BPT is an efficient tree-based data structure for model templates whose resolutions are coincided with those of an image pyramid.  It reduces the search space for 2D position and 3D pose simultaneously and make pose estimation faster.
 \item[\textbullet] Third component is optimum memory rearrangement for a coarse-to-fine search.  The input features at lower levels of an image pyramid are rearranged so that two types of features within 4-by-4 pixels are linearly aligned.  Template matching using SIMD instructions is efficiently executed on the rearranged feature map.
 \item[\textbullet] We created evaluation dataset in a bin-picking scene and make it publicly available \footnote{\,\url{http://isl.sist.chukyo-u.ac.jp/archives4/}}.  The dataset consists of six mechanical parts which are commonly seen in real factory lines.  The data were captured by an industrial 3D sensor and their 6D pose was fully annotated.
\end{itemize}

The proposed algorithm is evaluated on public tabletop dataset and our bin-picking dataset.  It is compared with the existing methods including recent CNN based approaches.

%%%%%%%%%%%%%%%%%%%%%%%%%%%%%%%%%%%%%%%%%%%%%%%%%%%%%%%%%%%%
%%%  Related Work    %%%%%%%%%%%%%%%%%%%%%%%%%%%%%%%%%%%%%%%
%%%%%%%%%%%%%%%%%%%%%%%%%%%%%%%%%%%%%%%%%%%%%%%%%%%%%%%%%%%%
\section{Related Work}
\label{sec:RelatedWork}
In this section, the existing researches on 6D object pose estimation are reviewed.  They are categorized into three approaches; template matching based, local descriptor based and learning based approaches.

\vspace{3pt} \noindent {\bf Template matching based approach.  }
The research on template matching based approach for 6D object pose estimation has started with monocular image in 1990s.  The whole appearance of a target object from various viewpoints were used as model templates and the matching between models and inputs was done based on line features \cite{Lanser1995}, edges and silhouettes \cite{Byne1998}, and shock graphs and curves \cite{Cyr2004}.

Adding depth information makes 6D object pose estimation more robust against background clutters.  The fast and robust RGB-D features such as VFH \cite{Rusu2010} and CVFH \cite{Aldoma2011} have been proposed for robotic applications.  Hinterstoisser et al. \cite{Hinterstoisser2012a, Hinterstoisser2012b} have proposed LINEMOD where discretized orientations of image gradients and surface normals were used as features.  The similarity scores were quickly computed on precomputed response maps and they showed that it was more robust against cluttered background and faster than the existing methods.

The coarse-to-fine search using image pyramid \cite{Borgefors1988} has been widely used for accelerating template matching.  This efficient search strategy in 2D image space has been applied to 6D object pose estimation by clustering similar templates \cite{Ulrich2012, Konishi2016}.  Moreover, template matching using hash tables \cite{Kehl2015} and GPU-optimized feature vectors \cite{Cao2016} have also been proposed for accelerating 6D pose estimation.

\vspace{3pt} \noindent {\bf Local descriptor based approach.  }
In local descriptor based approaches, 6D object pose is recovered from the correspondences or Hough voting based on local features.  Various local features extracted on 2D images such as line features \cite{David2005}, edges \cite{Choi2012b} and edgelet constellations \cite{Damen2012} have been proposed.  For more robust estimation, the local descriptors using depth information have also been proposed such as spin image \cite{Johnson1999} and SHOT \cite{Tombari2010}.

The point pair feature (PPF) \cite{Drost2010} is a most successful and well known 3D local descriptor ever and many extended versions have been proposed.  For example, selecting points of boundaries or lines \cite{Choi2012a}, calculating PPF on segmented point clouds \cite{Birdal2015} and modified point sampling and voting \cite{Hinterstoisser2016} have been proposed.  However, computing PPF for all pairs of input points is rather slow compared to template based approaches.

\vspace{3pt} \noindent {\bf Learning based approach.  }
Machine learning techniques have been utilized for extracting discriminative features and training classifiers which discriminate foreground/background, object classes and 3D object pose.  For example, learning the weights for template matching \cite{RiosCabrera2013} or voting \cite{Tuzel2014}, learning latent class distribution \cite{Tejani2014} and learned Hough forest for coordinate regression \cite{Brachmann2014} have been proposed.

In recent years, CNN has been introduced to learning the manifold of 3D object pose \cite{Wohlhart2015}.  The manifold learning based on convolutional auto encoder \cite{Kehl2016} and self-supervised augmented auto encoder \cite{Sundermeyer2018} have also been proposed.  Kehl et al. \cite{Kehl2017} have proposed SSD-like CNN architecture for estimation of 2D position, class and 3D pose of the object.  Instead of estimation of 3D pose class, the projected 2D points of 3D control points or bounding box corners were detected by CNN based detectors \cite{Crivellaro2015, Rad2017, Tekin2018}.

Although the recent CNN based approaches have often demonstrated higher robustness against background clutters and partial occlusions compared to other two approaches, their trainings require large number of annotated training samples and take longer hours on GPU.

\vspace{3pt} \noindent {\bf 6D object pose estimation in real applications.  }
The target objects are different in every real application such as robotic manipulation and AR.  It is too cumbersome to collect and annotate large number of training samples every time.  Therefore, template or local descriptor based approaches are reasonable because their models are trained solely from 3D CAD of the objects in a short time.  Regarding these two approaches, the existing researches have shown that the template matching based approaches were faster but less scalable to increasing number of object class than the local descriptor based approaches.

%%%%%%%%%%%%%%%%%%%%%%%%%%%%%%%%%%%%%%%%%%%%%%%%%%%%%%%%%%%%
%%%  Proposed Method  %%%%%%%%%%%%%%%%%%%%%%%%%%%%%%%%%%%%%%
%%%%%%%%%%%%%%%%%%%%%%%%%%%%%%%%%%%%%%%%%%%%%%%%%%%%%%%%%%%%
\section{Proposed Method}
\label{sec:ProposedMedhod}
This section introduces our proposed method which consists of three technical components: PCOF-MOD, BPT and memory rearrangement for a coarse-to-fine search.  We explain them in the following three subsections and the whole pipeline of 6D pose estimation in the last subsection.

%%========================================================%%
%%========================================================%%
\subsection{PCOF-MOD: multi-modal PCOF}
\label{subsec:PCOF-MOD}
PCOF-MOD is developed from PCOF (Perspectively Cumulated Orientation Feature) \cite{Konishi2016} which is robust to the appearance changes caused by the changes in 3D object pose.  PCOF is based on gradient orientaions extracted from RGB images and it represents the shapes of object contours.  Similar to LINEMOD \cite{Hinterstoisser2012a}, we add the orientation of surface normals extracted from depth images which represents the shapes of object surfaces.

\begin{figure}[tb]
 \begin{center}
  \begin{tabular}{c}
   \begin{minipage}{3.6cm}
    \begin{center}
     \includegraphics[clip, width=3.6cm]{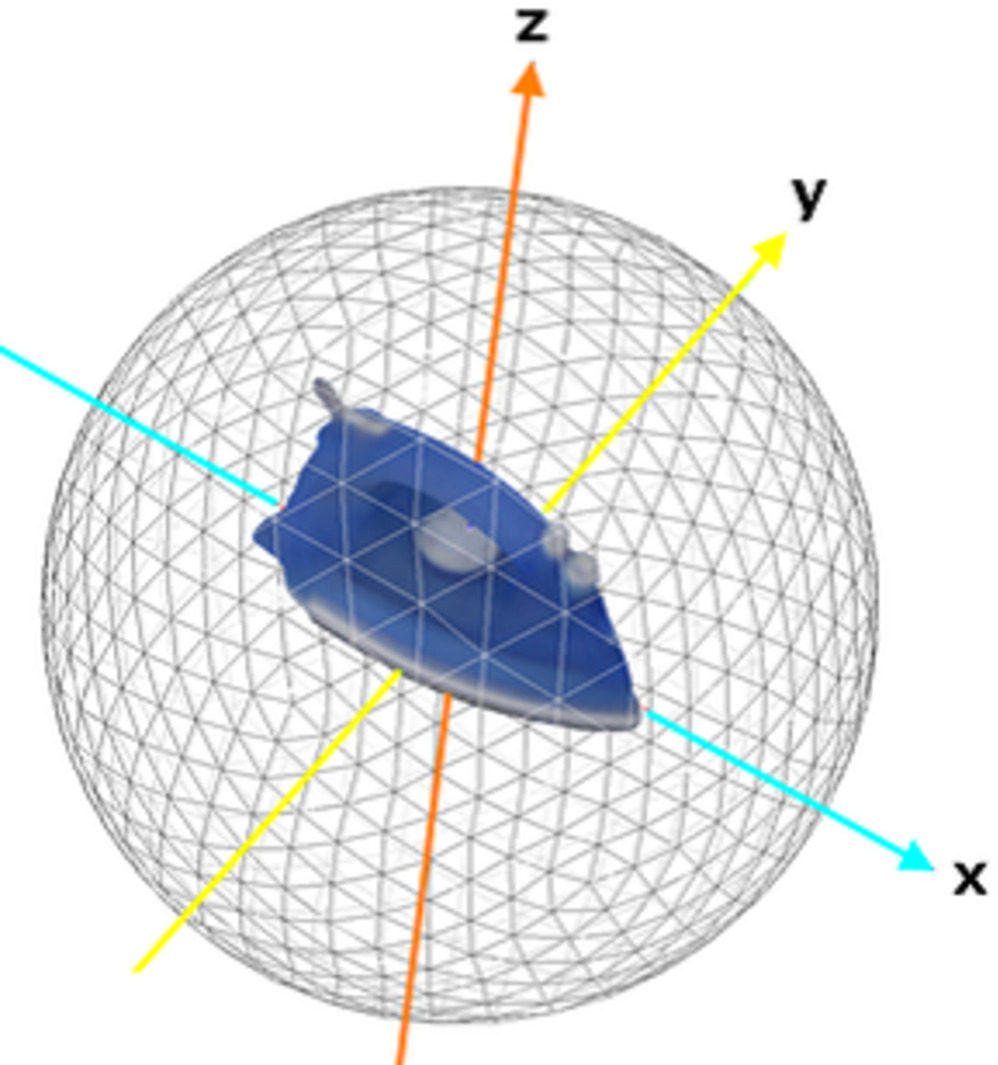}
     (a)
    \end{center}
   \end{minipage}
  \end{tabular}
  \\
  \begin{tabular}{c}
   \begin{minipage}{6.5cm}
    \begin{center}
     \includegraphics[clip, width=6.5cm]{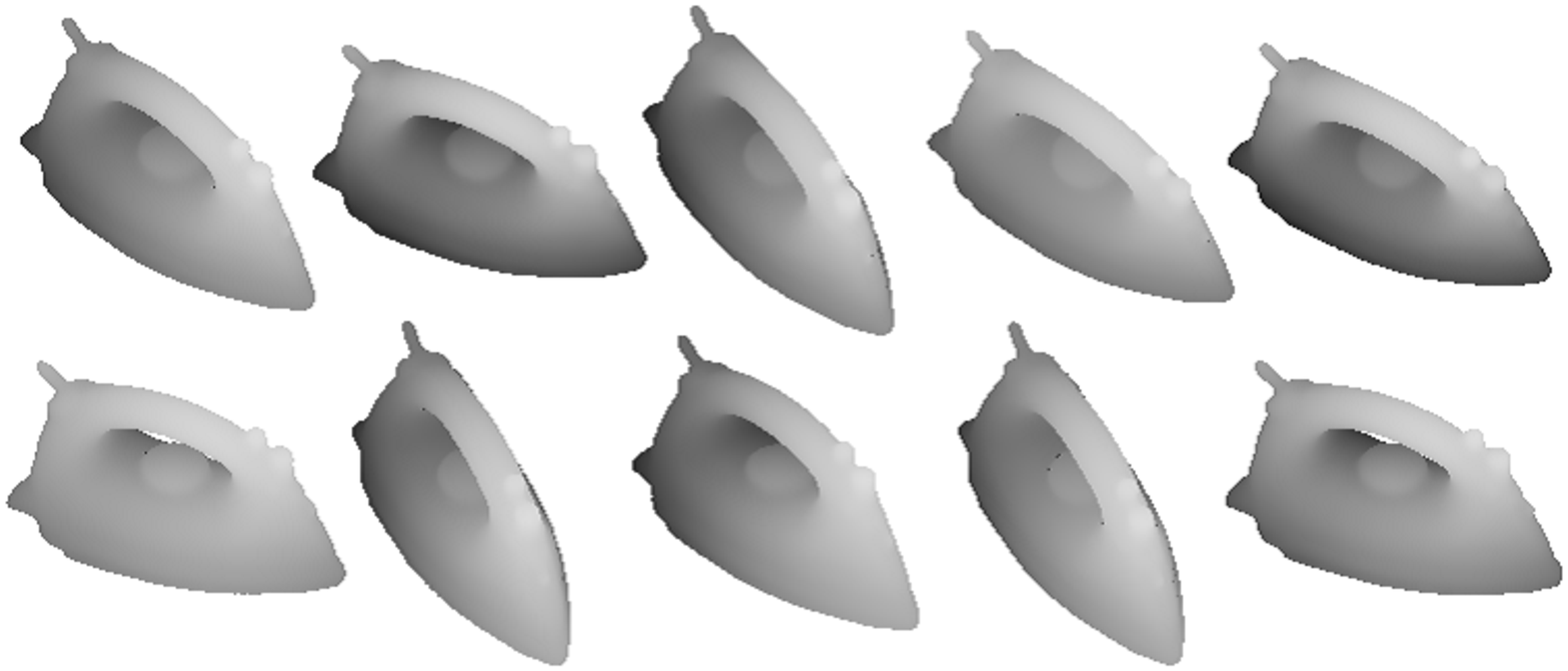}
     (b)
    \end{center}
   \end{minipage}
  \end{tabular}
 \end{center}
 \caption{(a) 3D CAD of iron, its coordinate axes and a sphere for viewpoint sampling.  (b) Examples of depth images from randomized viewpoints around a certain vertex.}
 \label{fig:CAD_image}
\end{figure}

We describe the details of PCOF-MOD using CAD of iron (Figure \ref{fig:CAD_image}(a)) in ACCV dataset (Subsection \ref{subsec:ex1}) as an example.  Firstly, depth images are rendered from randomized viewpoints sampled on the spheres whose coordinate axes are aligned with those of target objects.  4 parameters which determine the viewpoints (rotation angles around x, y, optical axes, distance from the object) are generated using uniform random number in a certain range.  This range of randomization should be small enough for a single template at a viewpoint can represent the distributions of features.  In our research, we experimentally determined the ranges and they are $\pm 10$ degrees around xy axes, $\pm 7.5$ degrees around optical axis and $\pm 90$ mm from objects.  Internal camera parameters for rendering depth images should be same as ones of the RGB-D sensor used for pose estimation.  Some of the depth images rendered using randomized viewpoints (the center of the range is at 33.9 deg around x axis, 25.5 deg around y axis 0 deg around optical axis and 900 mm from the object) are shown in Figure \ref{fig:CAD_image}(b).  The upper left image is rendered at the center of the randomization range.

Secondly, gradient vectors and normal vectors are extracted from the rendered $N$ depth images.  The gradient vectors are computed only around object contours using Sobel filter and the normal vectors are computed by fitting a plane to nearby pixels.

Thirdly, the distributions of the gradient and normal orientations are computed at each pixel.  The directions of gradient and normal vectors are quantized into eight orientations and weights are added to corresponding bins.  The weights are linearly interpolated between neighboring bins and added to them.  When there is no depth value, no weight is added to the histogram at the pixel.  Then two histograms are obtained per pixel whose maximum frequencies are $N$.

Lastly, we select dominant orientations whose frequencies are larger than a certain threshold ($Th$) and extract 8 bit binary digits where the bit of the dominant orientations are 1.  The frequency values of the maximum bin are also extracted and used as the weight for calculating similarity scores because the features with higher frequencies are more stably observed and more robust against the changes of 3D object pose.  The histograms without the dominant orientation are not used for calculating the score.

\begin{figure}[tb]
 \begin{center}
  \includegraphics[width=8.2cm]{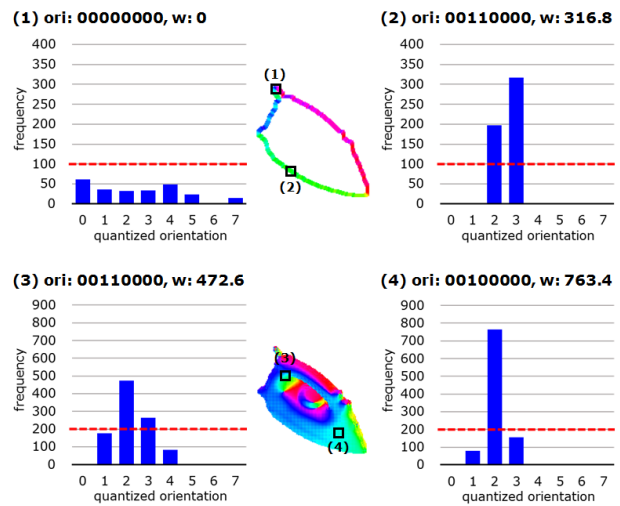}
 \end{center}
 \caption{Examples of the orientation histograms, binary features (ori) and their weights (w) on arbitrarily selected pixels.  Pixel A and B are extracted from gradient orientations, and pixel C and D are from normal orientations.  Red dotted lines show the threshold for feature extraction}
 \label{fig:histogram}
\end{figure}

Four examples of histograms, quantized orientation features (ori) and weights (w) are shown in Figure \ref{fig:histogram}.  These are calculated from the depth images shown in Figure \ref{fig:CAD_image}(b).  The pixel A and B are selected from the gradient orientation image and the pixel C and D are from the normal orientation image.  The number of generated depth images ($N$) and the threshold for frequencies ($Th$) were experimentally determined.   We used $N = 1000$ and $Th = 100$ for the gradient orientations and $N = 1000$ and $Th = 200$ for the normal orientations.  Regarding the gradient orientations, the votes are distributed to many bins (orientations) and the dominant orientations are not obtained on the corners of objects like pixel A.  Contrastingly, the votes are concentrated on a few bins and the dominant orientations with large weights are obtained on the smooth contours of objects like pixel B.  Similar to the gradient orientations, the normal orientations with smaller weights are extracted on the corner shapes like pixel C and the orientations with larger weights are extracted on the smooth surface like pixel D.

Two kinds of templates ($Tg$: gradient orientation templates, $Tn$: normal orientation templates) are created at each viewpoint and each template consists of $n$ quantized orientations ($ori_i$) and weights ($w_i$) of pixels ($x_i$ and $y_i$) whose weights are larger than zero: 
\begin{eqnarray}
T: \left\{ x_i, y_i, ori_i, w_i | i = 1, ... , n \right\}.
\label{eqn:template}
\end{eqnarray}
A similarity score at pixel (x, y) is calculated by following equations for each template: 
\begin{eqnarray}
score(x,y) = \frac{\sum^{n}_{i=1} \delta_k(ori^{I}_{(x+x_i,y+y_i)} \in ori^{T}_{i})}{\sum^n_{i=1} w_i}.
\label{eqn:score}
\end{eqnarray}

The weights are added to the score when any of the orientations of an input image are included in the orientations of model template.  The delta function in Equation \ref{eqn:score} can be computed efficiently using bitwise AND ($\land$).

\begin{eqnarray}
\delta_i(ori^{I} \in ori^{T}) = 
\begin{cases}
w_i & \text{if $ori^{I} \land ori^{T} > 0$}, \\
0   & \text{otherwise}.
\end{cases}
\label{eqn:matching}
\end{eqnarray}

%%========================================================%%
%%========================================================%%
\subsection{BPT: Balanced Pose Tree}
\label{subsec:BPT}
The efficient tree-based data structures have been proposed for 6D object pose estimation from a RGB image \cite{Ulrich2012, Konishi2016}.  They clustered 3D viewpoints based solely on 2D view similarities.  However, the number of child nodes sometimes becomes large because a simple-shaped object has many similar 2D views, and this imbalanced trees lead to slower computations.  We propose new hierarchical template structure which is based on regularly sampled viewpoints where the numbers of child nodes of all parent nodes are almost the same (balanced tree).

\begin{figure}[tb]
 \begin{center}
  \begin{tabular}{c}
   \begin{minipage}{1.6cm}
    \begin{center}
     \includegraphics[clip, width=1.6cm]{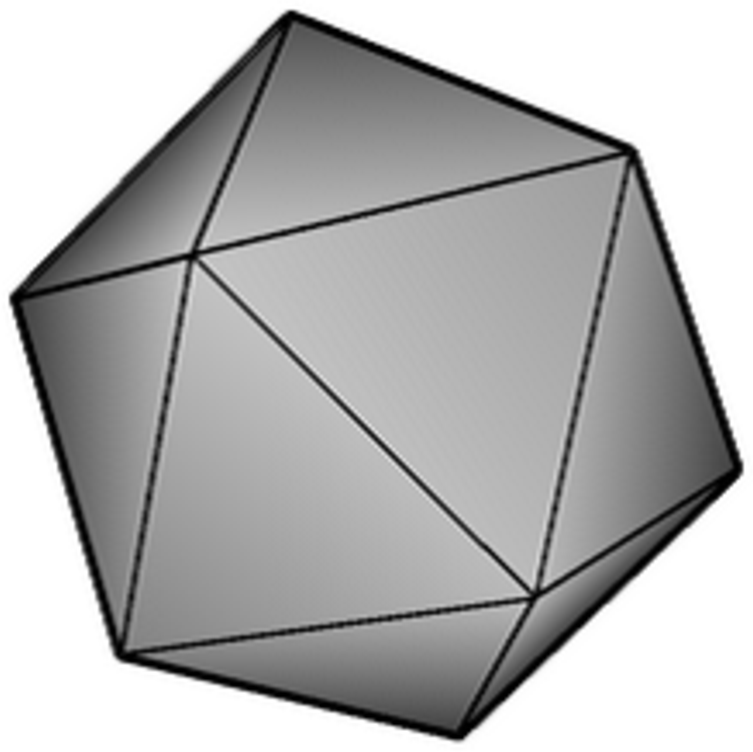}
     depth 0
    \end{center}
   \end{minipage}
   \hspace{0.2cm}
   \begin{minipage}{1.6cm}
    \begin{center}
     \includegraphics[clip, width=1.6cm]{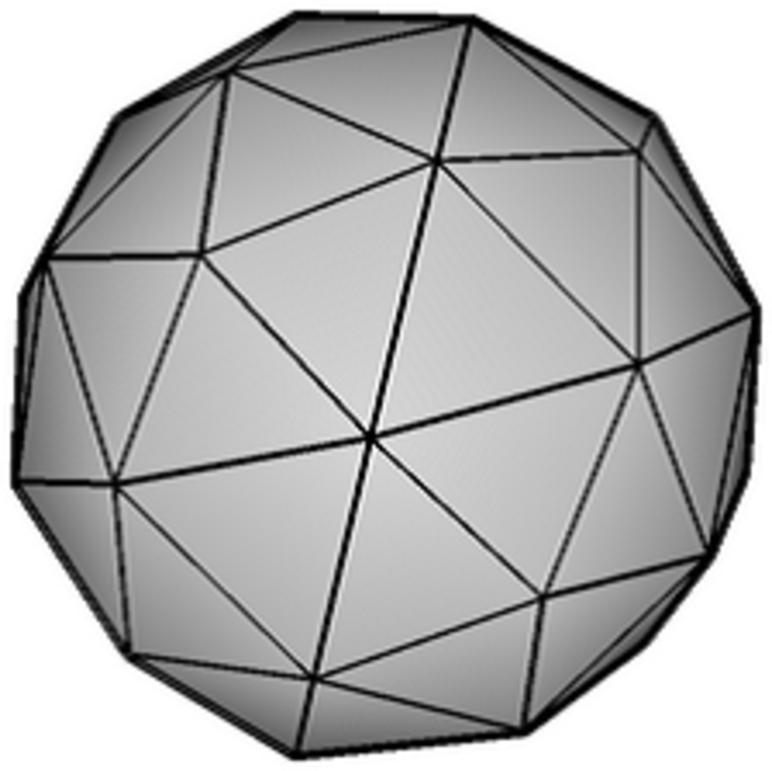}
     depth 1
    \end{center}
   \end{minipage}
   \hspace{0.2cm}
   \begin{minipage}{1.6cm}
    \begin{center}
     \includegraphics[clip, width=1.6cm]{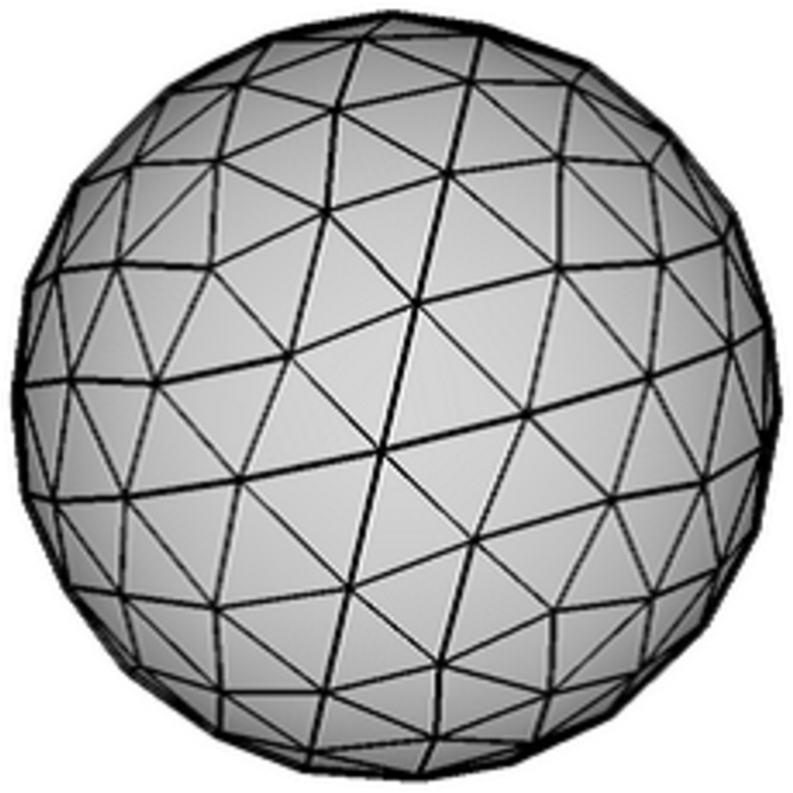}
     depth 2
    \end{center}
   \end{minipage}
   \hspace{0.2cm}
   \begin{minipage}{1.6cm}
    \begin{center}
     \includegraphics[clip, width=1.6cm]{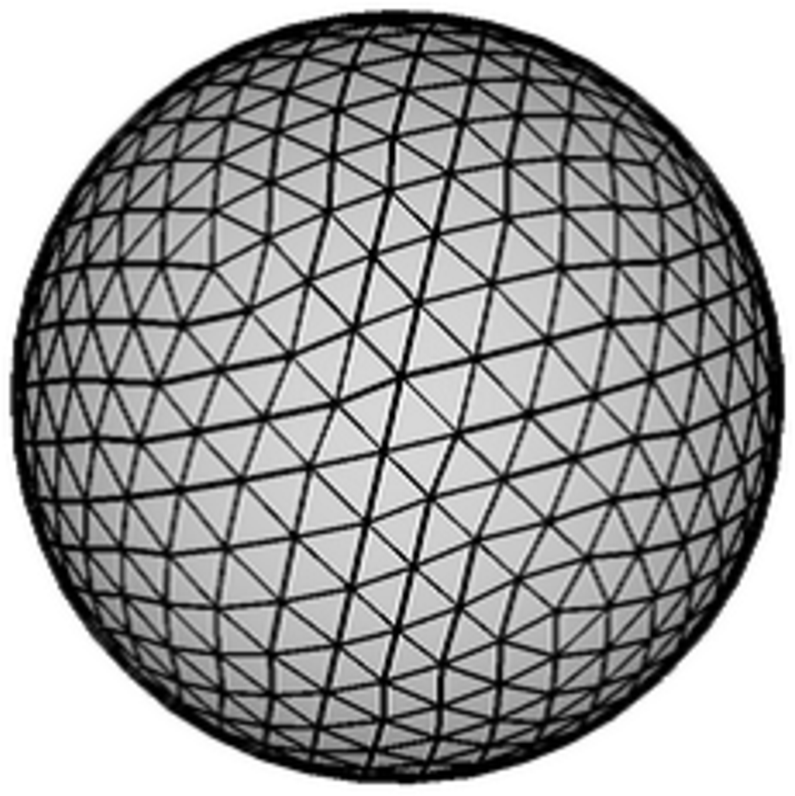}
     depth 3
    \end{center}
   \end{minipage}
  \end{tabular}
 \end{center}
 \caption{Icosahedron (left) and almost regular polyhedrons those are generated by recursive decompositions.}
 \label{fig:sphere}
\end{figure}

The PCOF-MOD template described in previous subsection is robust to small change in object 3D pose, e.g. within $\pm 10$ degrees around xy axes, $\pm 7.5$ degrees around optical axis and $\pm 90$ mm of the distance to an object in our research.  To cover full 3D object pose, PCOF-MOD templates are made at the viewpoints which are regularly sampled on the sphere (Figure \ref{fig:CAD_image} (a)).  These viewpoints are the vertices of an almost regular polyhedron and are made by recursively decomposing an icosahedron.  Figure \ref{fig:sphere} shows this procedure where new vertices are made by dividing edges in half.  It starts from the icosahedron (20 faces) shown in leftmost of Figure \ref{fig:sphere} and the polyhedrons with 80 faces, 320 faces and 1280 faces are obtained in sequence.  The number of vertices (viewpoints) are 12, 42, 162 and 642 respectively.

We used the 1280 faced polyhedrons (642 vertices) for sampling viewpoints of PCOF-MOD templates because the angles between neighboring viewpoints are approximately 8 degrees around xy axes and single PCOF-MOD template ($\pm 10$ degrees) can fully cover this range.  The templates are also made at 70 mm intervals in distance to an object and 6 degrees intervals around an optical axis so that it ($\pm 90$ mm and $\pm 7.5$ degrees) can fully cover these intervals.

We integrate all these templates into balanced pose tree (BPT) which consists of the hierarchical templates with different resolutions and viewpoint intervals.  We used sparse viewpoints at the lower image resolutions because the small differences in 3D pose are not recognized at the lower resolution images.  This reduces the number of templates to be scanned at the higher pyramid levels and make pose estimation more efficient.

Our BPT consists of four levels (depth 0 ... 3) and the viewpoint sampling becomes denser at the deeper layer.  We use the vertices of icosahedron shown in the left of Figure \ref{fig:sphere} as the root nodes of BPT and link each root node to its nearest vertices of depth 1 (80 faced polyhedron).  Each parent node has three or four child nodes and this procedure is iterated from depth 1 to depth 2 and from depth 2 to depth 3.  We also decrease the intervals by half for the rotation angles around optical axis and the distance to the object.  Therefore, our BPT is a group of B-trees of depth 3 where each parent node has 12 or 16 child nodes.

\begin{algorithm}[tb]
\caption{Building balanced pose tree}
\label{alg:BPT}
\begin{algorithmic}
\REQUIRE Orientation histograms $H_3$ at level 3
\ENSURE Templates $T_i$ ($i = 0, ..., d-1$)
\FOR {$i \leftarrow d-1 $ to $0$}
  \STATE $ P_i \leftarrow $ parent viewpoints of $i$th level in $BPT$
  \FOR {each parent viewpoint $ P_{i,j} $}
    \STATE $ C_{i+1,j} \leftarrow $ child viewpoints of $P_{i,j}$
    \STATE $ H'_{i+1,j} \leftarrow $ add histograms of child viewpoints $ H_{i+1} \in C_{i+1,j} $ at each pixel
    \STATE $ H^{''}_{i+1,j} \leftarrow $ normalize histograms $H'_{i+1,j}$
    \STATE $ H_{i,j} \leftarrow $ add histograms of nearby $2 \times 2$ px of $H^{''}_{i+1,j}$
    \STATE $ H'_{i,j} \leftarrow $ normalize histograms $H_{i,j}$
    \STATE $ T_{i,j} \leftarrow $ thresholding $H'_{i,j}$ and extracting new binary features and weights
  \ENDFOR
\ENDFOR
\end{algorithmic}
\end{algorithm}

The gradient and normal templates ($Tg$ and $Tn$) at depth 2 and the upper levels are made in a bottom-up way using the templates of one level lower and the algorithm is shown in Algorithm \ref{alg:BPT}.  Firstly, the orientation histograms ($H_{i+1}$) of the child nodes ($C_{i+1}$) which has same parent node ($P_i$) are added and normalized at each pixel.  The number of child nodes are 12 or 16 and their 3D pose (including the angles around optical axis and the distance to the object) are slightly different.  The added histograms represent wider distribution of orientation which should be handled by the parent node.  Secondly, the resolution of the added histograms ($H^{''}_{i+1,j}$) are reduced to half by adding and normalizing the histograms of nearby $2 \times 2$ pixels.  Lastly, the binary orientation features and weights of the templates ($T_{i,j}$) are extracted by thresholding the histograms.  These procedures are iterated to depth 0.

%%========================================================%%
%%========================================================%%
\subsection{Memory rearrangement for a coarse-to-fine search}
\label{subsec:MemoryRearrangement}
The memory rearrangement techniques have been proposed for fast template matching of spreading features \cite{Hinterstoisser2012a} and on GPU \cite{Cao2016}.  We are inspired by these researches and propose novel rearrangement algorithm which boosts a coarse-to-fine search with an image pyramid.

\begin{figure}[tb]
\begin{center}
\includegraphics[width=8.2cm]{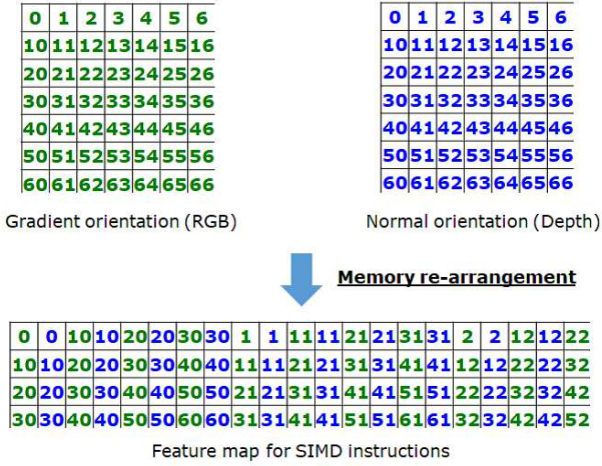}
\end{center}
\caption{Our memory rearrangement algorithm which enables a highly efficient coarse-to-fine search.  The upper two figures show the gradient orientation features (green) and normal orientation features (blue) at the lower levels of an image pyramid.  The colored numbers indicate the memory address.  These two features are mixed and rearranged so that every 4 by 4 grid of these features are linearly aligned (the lower figure).}
\label{fig:feature_map}
\end{figure}

Our pose estimation algorithm uses two kinds of binary features, one is quantized gradient orientations extracted from RGB image and another is quantized normal orientations extracted from depth image.  The upper images of Figure \ref{fig:feature_map} show the part of features extracted from 10-by-10 pixel size images.  The blue and green numbers indicate the memory address which start at the top-left of the images.

When the template is scanned exhaustively on the top of the image pyramid, the calculation of similarity scores is easily accelerated by using SIMD instructions.  In case of Intel AVX intrinsics, 256 bit register is available and 32 model features (8 bit) are compared with the input features by single instruction (logical AND in Equation \ref{eqn:matching}).

However, on the lower levels of the pyramid, the templates are searched only around the promising areas selected by the matching results of one level upper.  On the image pyramids where the size of the lower image is increased to double, the templates are searched in 2-by-2 pixels, for example '0', '1', '10', '11' in Figure \ref{fig:feature_map}.  In this case, the features to be compared with the templates are not linearly aligned and applying SIMD instructions is not effective.  We propose the algorithm which rearranges nearby features in a rectangular grid into a linearly aligned form and the template matching is done highly efficiently on the rearranged feature map using SIMD instructions.

In this paper, two kinds of 8 bit features are used and 32 features are processed at one time using Intel AVX intrinsics.  In order to make full use of this, we rearrange these two kinds of features in 4-by-4 pixels into 32 linearly aligned features (the lower image in Figure \ref{fig:feature_map}).  When there is promising results in 2-by-2 pixels at the upper level of the pyramids, the corresponding 4-by-4 pixels at the lower level are searched and any 4-by-4 features on the rearranged feature map can be accessed in a linearly aligned form.  The model templates (Equation \ref{eqn:template}) also should be rearranged in the same order as the feature map.

Our proposed feature rearrangement for an efficient coarse-to-fine search can be applied to any length binary or floating-point features.  We should note that the rearrangement takes a few milliseconds and the rearranged feature map consumes more memory by 4 times than original input features.

%%========================================================%%
%%========================================================%%
\subsection{6D object pose estimation and refinement}
\label{subsec:PoseEstimation}
In pose estimation, firstly, the image pyramids of RGB-D input are made and the quantized gradient and normal orientations are computed from a RGB and depth image.  Secondly, the root nodes of BPT are scanned at the top level of the feature pyramids.  The similarity scores of the gradients and the normals are computed using Equation \ref{eqn:score} and the results whose sum of the scores are larger than a certain threshold are selected as the promising results.  These results (pose and position) are further searched at the lower levels of the pyramids (on the rearranged feature map) using the templates at the lower depth of the tree.  At the bottom of the pyramids, the detected positions on the image and their 3D pose of the matched templates are obtained after non-maximum suppression.  The correspondences between 2D points on the image and 3D points on CAD are obtained and 6D object pose is retrieved by solving P$n$P problem.  Lastly, the obtained 6D pose is refined using ICP algorithm.

%%%%%%%%%%%%%%%%%%%%%%%%%%%%%%%%%%%%%%%%%%%%%%%%%%%%%%%%%%%%
%%%  5.3. Experimental Evaluation  %%%%%%%%%%%%%%%%%%%%%%%%%
%%%%%%%%%%%%%%%%%%%%%%%%%%%%%%%%%%%%%%%%%%%%%%%%%%%%%%%%%%%%
\begin{figure*}[tb]
 \begin{center}
  \begin{tabular}{c}
   \begin{minipage}{17cm}
    \begin{center}
     \includegraphics[clip, width=3.9cm]{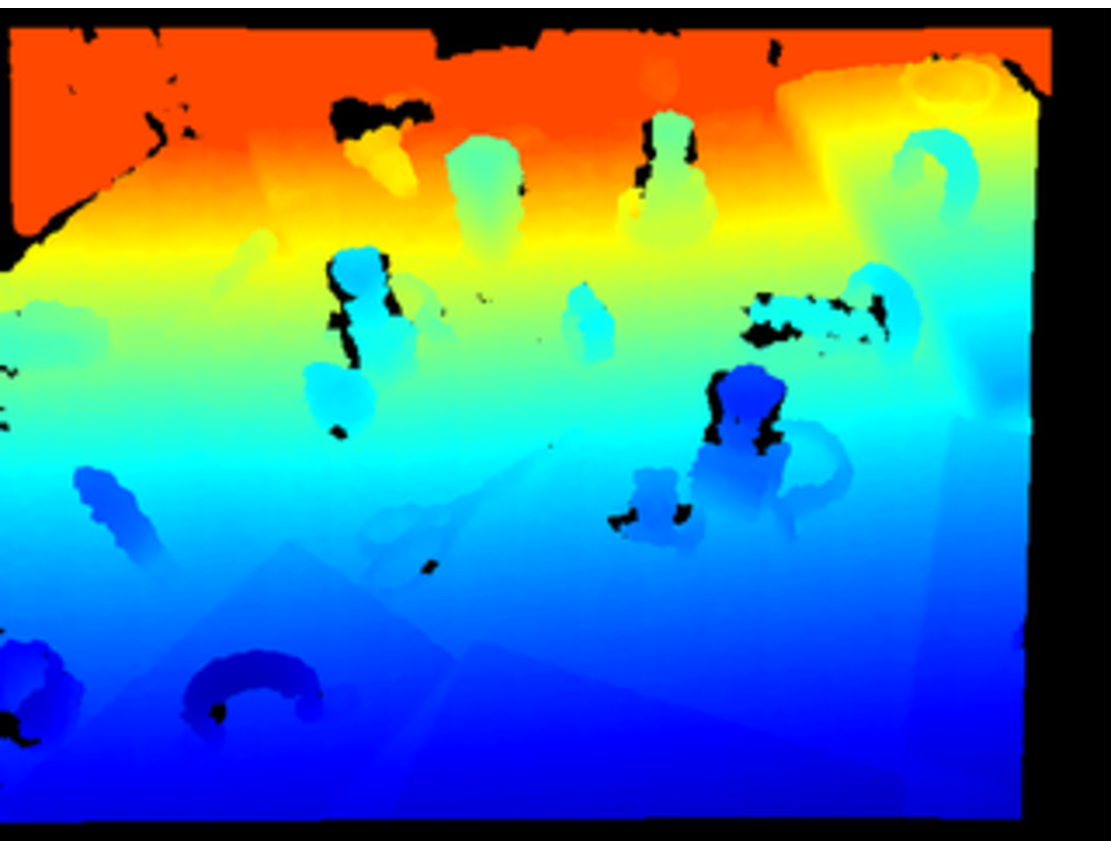}
     %%\hspace{0.1cm}
     \includegraphics[clip, width=3.9cm]{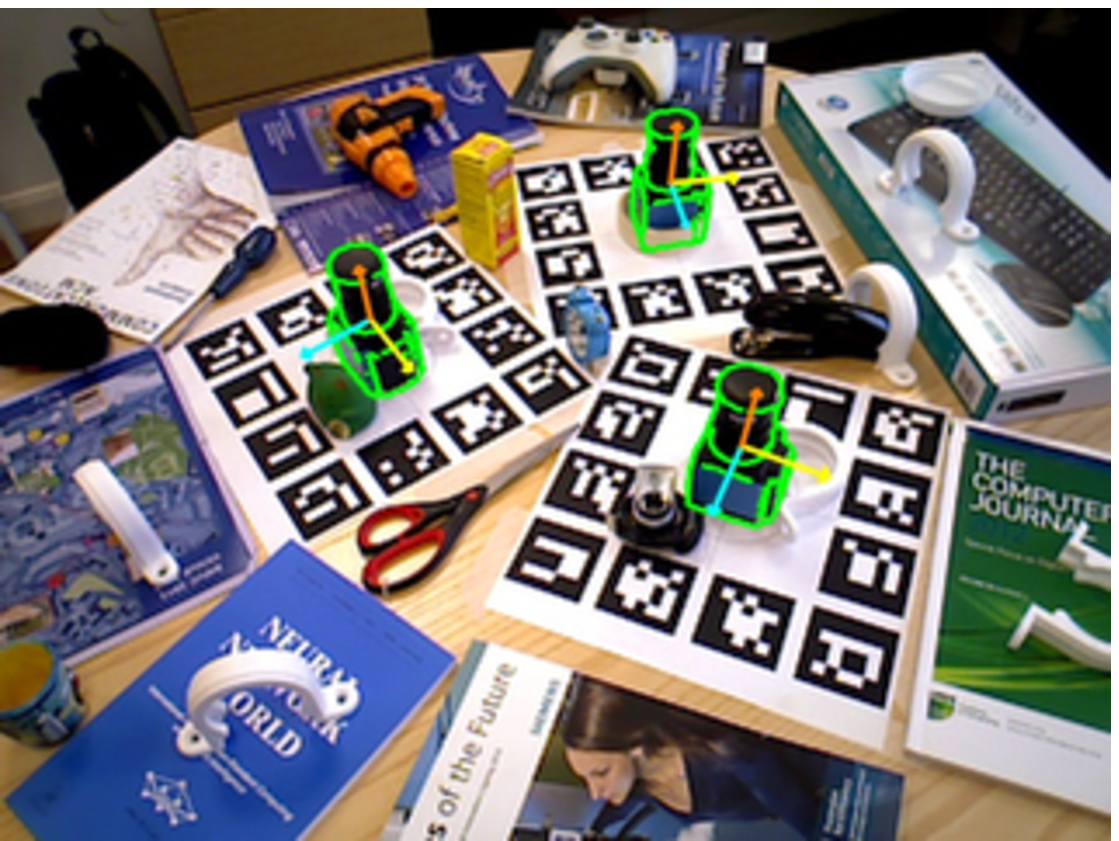}
     \hspace{0.3cm}
     \includegraphics[clip, width=3.9cm]{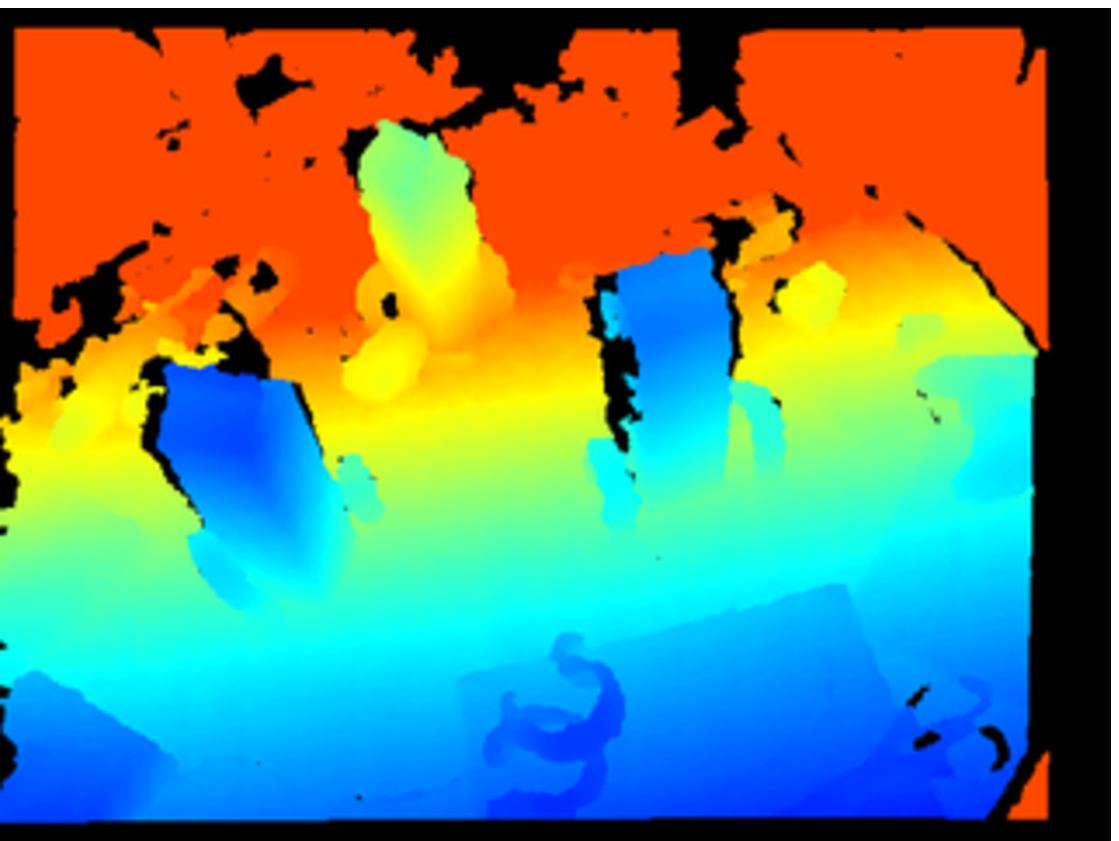}
     %%\hspace{0.1cm}
     \includegraphics[clip, width=3.9cm]{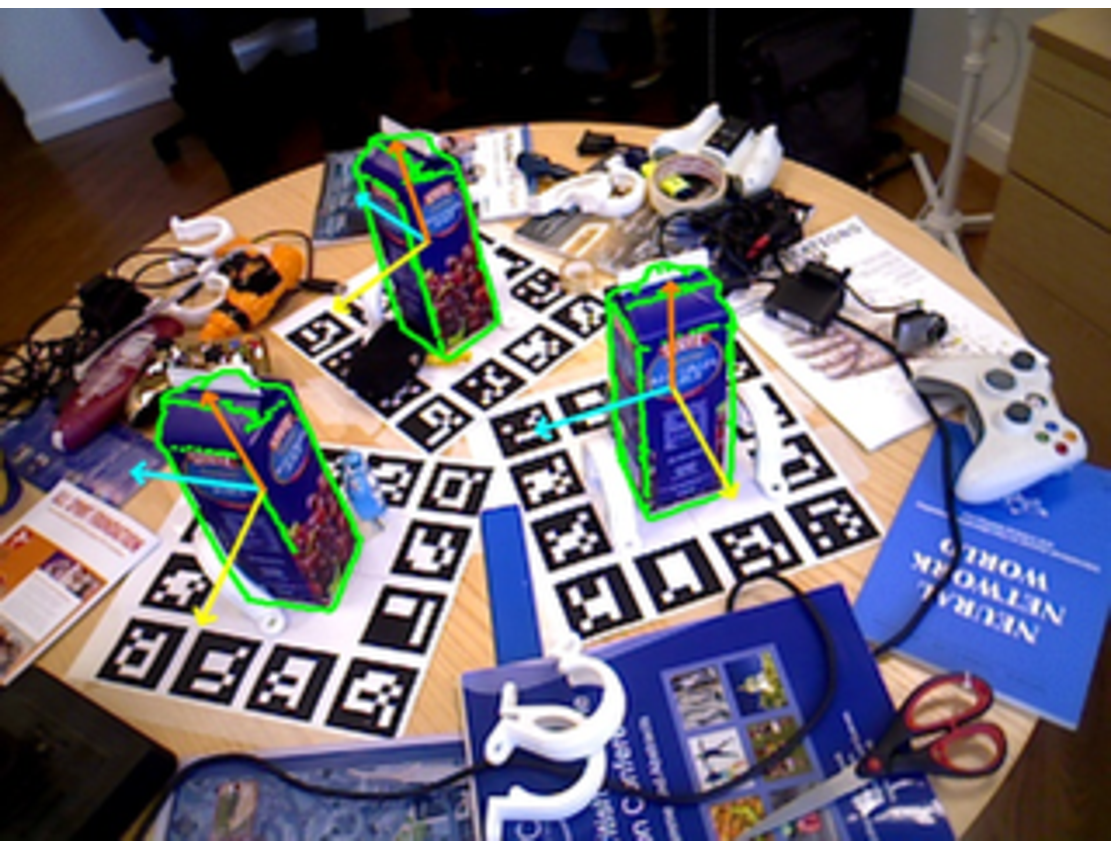}
    \end{center}
   \end{minipage}
  \end{tabular}
  \\
  \vspace{0.05cm}
  \begin{tabular}{c}
   \begin{minipage}{17cm}
    \begin{center}
     \includegraphics[clip, width=3.9cm]{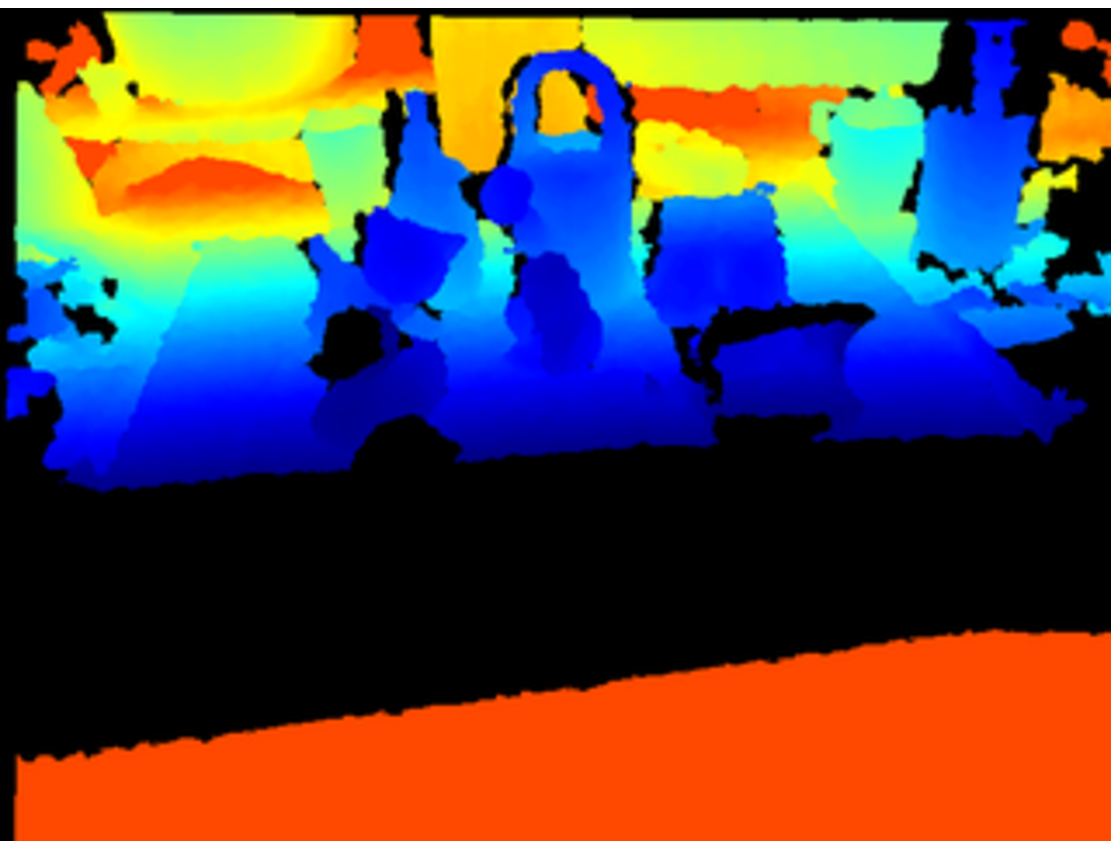}
     %%\hspace{0.1cm}
     \includegraphics[clip, width=3.9cm]{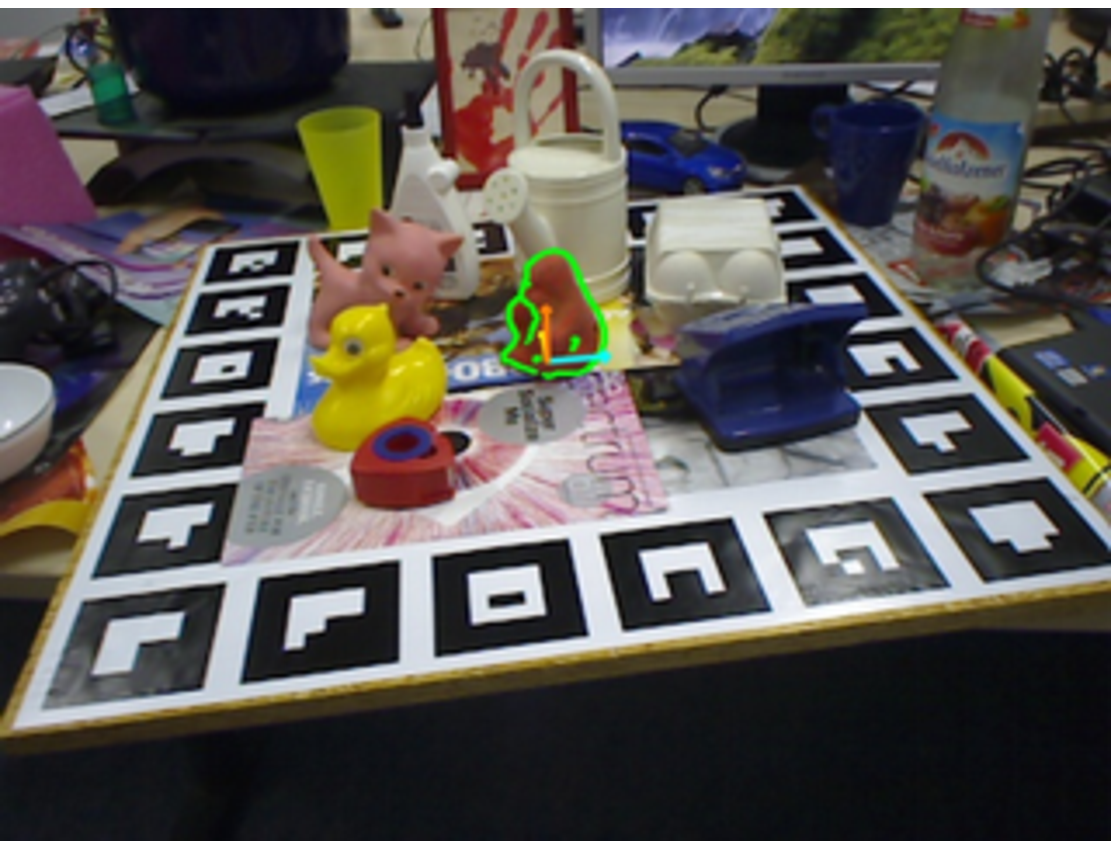}
     \hspace{0.3cm}
     \includegraphics[clip, width=3.9cm]{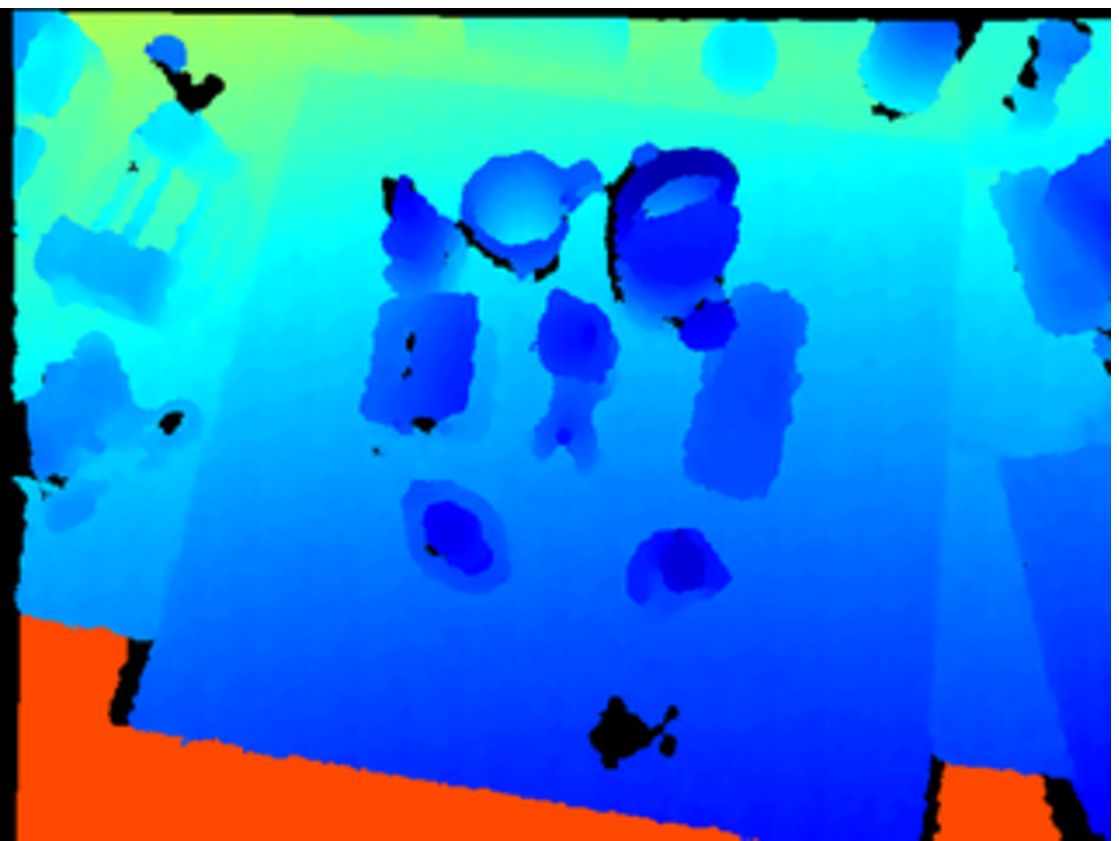}
     %%\hspace{0.1cm}
     \includegraphics[clip, width=3.9cm]{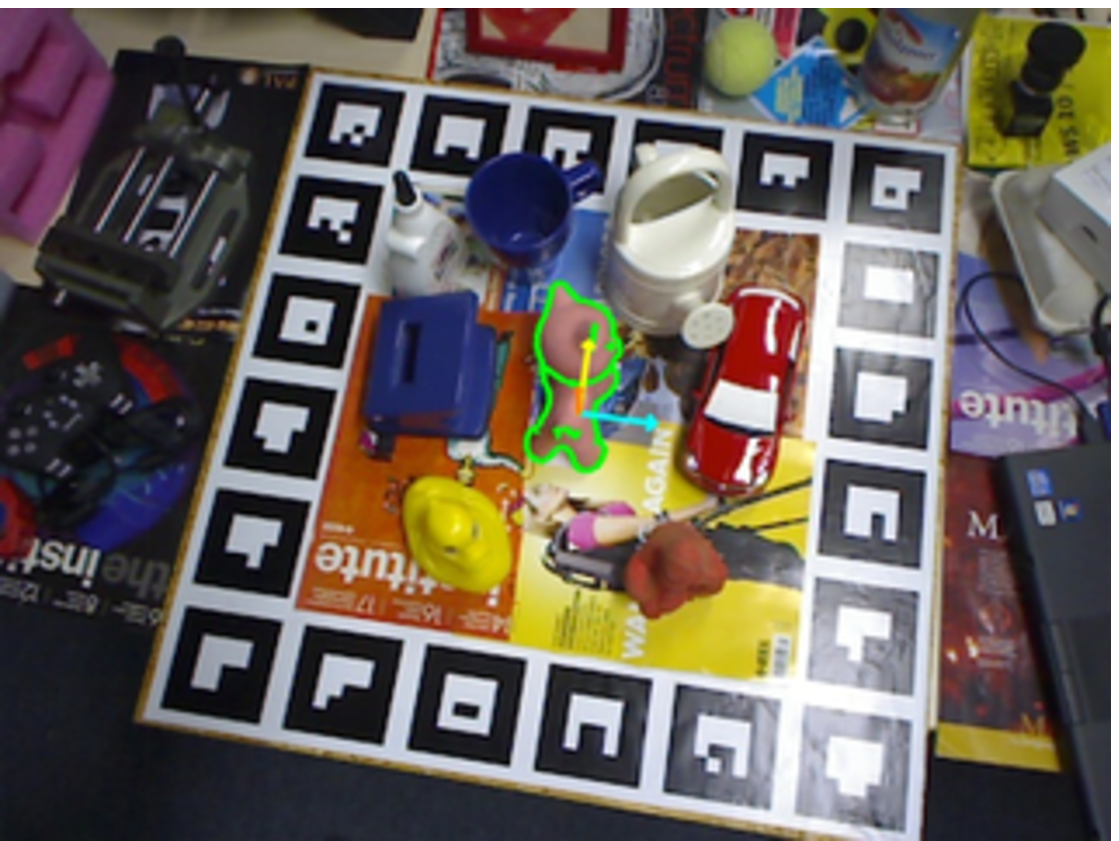}
    \end{center}
   \end{minipage}
  \end{tabular}
 \end{center}
 \caption{Example images of ICVL and ACCV dataset (Top row: Camera and Juice from ICVL)CBottom row: Ape and Cat from ACCV).  The depth and RGB images are shown for each object.  The edges of the objects extracted from 3D CAD (green lines) and the coordinate axes (three colored arrows) are drawn on the RGB images based on the estimated 6D pose by our proposed method.}
 \label{fig:ex1_image}
\end{figure*}

\section{Experimental Evaluation}
\label{sec:Experiment}
We evaluated our proposed algorithm and compared it with state-of-the-art methods in different two scenes.  One was tabletop scene for AR and service robots in housing space.  Another was bin-picking scene for industrial robots in factories.

%%========================================================%%
%%========================================================%%
\subsection{Evaluation on public tabletop dataset}
\label{subsec:ex1}
For evaluation in a tabletop scene, ICVL dataset \cite{Tejani2014} (we used corrected annotation \cite{Kehl2016}) and ACCV dataset \cite{Hinterstoisser2012b} which were publicly available RGB-D dataset were used.  ICVL dataset consists of 6 kinds of objects and ACCV dataset consists of 15 kinds of objects (13 objects whose CAD are provided were evaluated).  Both of them provide CAD data of the target objects and the ground truth of 6D object pose which were annotated using AR markers.  Examples of depth and RGB images of both dataset are shown in Figure \ref{fig:ex1_image}.

The pose of objects in both dataset ranges from -90 to 90 degrees around x and y axes, from -45 to 45 degrees around z axis (optical axis).  The distance from the camera ranges from 450 to 1100 mm for ICVL dataset and from 650 to 1150 mm for ACCV dataset.  For our proposed algorithm, model templates were trained in the ranges of the object pose.  Our algorithm was implemented using C++ and ran on Windows PC (Core i7-7700 3.6GHz) using 4 cores.  The estimation accuracy and speed was compared with the existing template based \cite{Hinterstoisser2012b} and learning based \cite{Tejani2014, Kehl2016, Kehl2017} methods.

We ran our program using various thresholds and calculated recall, precision and F1 score.  When the average distance between the model points which are transformed by ground truth and those by estimated pose is smaller than $k_md$, the estimation is regarded as correct. $d$ is the diameter of the object and we use 0.15 as the coefficient $k_m$ which is same in \cite{Kehl2016}.

%%--------------------------------------------------------%%
\begin{table}[tb]
 \caption{F1 scores on ICVL dataset.}
 \begin{center}
   {\tabcolsep = 2.5mm
    \begin{tabular}{l|c|c|c|c|c}
    \hline
             & LINEMOD       & LC-HF &       Kehl &        SSD-6D & \textbf{Ours} \\
    \hline
    Camera   &         0.589 & 0.394 &      0.383 & \textbf{0.741}&         0.627 \\
    Cup      &         0.942 & 0.891 &      0.972 &         0.983 & \textbf{0.992}\\
    Joystick &         0.846 & 0.549 &      0.892 & \textbf{0.997}&         0.975 \\
    Juice    &         0.595 & 0.883 &      0.866 &         0.919 & \textbf{0.945}\\
    Milk     &         0.558 & 0.397 &      0.463 & \textbf{0.780}&         0.719 \\
    Shampoo  & \textbf{0.922}& 0.792 &      0.910 &         0.892 &         0.897 \\
    \hline
    Mean     &         0.740 & 0.651 &      0.747 & \textbf{0.885}&         0.859 \\
    \hline
   \end{tabular}
 \label{tbl:ex1_ICVL_F1}
 }
 \end{center}
\end{table}

\begin{table}[tb]
 \caption{F1 scores on ACCV dataset.}
 \begin{center}
   {\tabcolsep = 2.5mm
    \begin{tabular}{l|c|c|c|c|c}
    \hline
    & LINEMOD & LC-HF & Kehl & SSD-6D & \textbf{Ours} \\
    \hline
    Ape     &   0.533 &     0.855 & \textbf{0.981}&         0.763 &         0.913 \\
    Vise    &   0.846 &     0.961 &         0.948 &         0.971 & \textbf{0.998}\\
    Cam     &   0.640 &     0.718 &         0.934 & \textbf{0.922}& \textbf{0.995}\\
    Can     &   0.512 &     0.709 &         0.826 &         0.931 & \textbf{0.988}\\
    Cat     &   0.656 &     0.888 &         0.981 & \textbf{0.893}& \textbf{0.997}\\
    Driller &   0.691 &     0.905 &         0.965 & \textbf{0.978}&         0.947 \\
    Duck    &   0.580 &     0.907 &         0.979 &         0.800 & \textbf{0.996}\\
    Eggbox  &   0.860 &     0.740 & \textbf{1.000}&         0.936 &         0.995 \\
    Glue    &   0.438 &     0.678 &         0.741 &         0.763 & \textbf{0.926}\\
    Puncher &   0.516 &     0.875 & \textbf{0.979}&         0.716 &         0.976 \\
    Iron    &   0.683 &     0.735 &         0.910 &         0.982 & \textbf{0.995}\\
    Lamp    &   0.675 &     0.921 & \textbf{0.982}&         0.930 &         0.914 \\
    Phone   &   0.563 &     0.728 &         0.849 &         0.924 & \textbf{0.992}\\
    \hline
    Mean    &   0.630 &     0.817 &         0.929 &         0.885 & \textbf{0.972}\\
    \hline
   \end{tabular}
 \label{tbl:ex1_ACCV3D_F1}
  }
 \end{center}
\end{table}

\vspace{3pt} \noindent {\bf Estimation accuracy.  }
The F1 scores on ICVL and ACCV dataset are shown in Table \ref{tbl:ex1_ICVL_F1} and Table \ref{tbl:ex1_ACCV3D_F1}.  The F1 scores by existing algorithms which were evaluated in \cite{Kehl2016, Kehl2017} are also shown in the tables.  Our proposed algorithm achieved the highest score among the state-of-the-art methods on ACCV dataset and the second highest on ICVL dataset.

There are two reasons why our algorithm has an advantage over existing template based (LINEMOD) and learning based (LC-HF and Kehl) algorithms.  One is that we sample viewpoints, roll angles and camera distances more densely for making model templates.  On ACCV dataset, our algorithm uses 41,088 templates (321 viewpoints, 16 roll angles and 8 distances) and the existing algorithms use 3,402 templates (81 viewpoints, 7 roll angles and 6 distances).  Using more templates makes pose estimation more accurate and robust because the pose differences between objects in captured images and model templates become smaller.  Another reason is that our proposed feature PCOF-MOD is extracted from a large number of depth images those are synthesized within a certain range of 3D object pose and the feature is robust to small pose changes of objects.  Due to this, PCOF-MOD relaxes only the matching conditions for target objects without increasing false positives under cluttered background and is matched to the testing objects whose poses are slightly different from those of model templates.

Though SSD-6D is learning based method which discriminates target objects from background, it uses only a RGB image.  Hinterstoisser et al. \cite{Hinterstoisser2012a} showed that adding depth information made pose estimation more robust against background clutters.  This is why the mean F1 score of SSD-6D was lower than those of Kehl and ours which used both of RGB and depth on ACCV dataset.  Meanwhile, the mean F1 score of SSD-6D was highest on ICVL dataset.  This is due to that ICVL dataset includes more partial occlusions compared to ACCV dataset.  These occlusions have influenced both on RGB and depth features, and degrade more largely the performance of RGB-D based methods.

%%--------------------------------------------------------%%
\begin{table}[tb]
 \caption{Processing time (ms) on ACCV dataset.}
 \begin{center}
  {\tabcolsep = 3mm
   \begin{tabular}{l|c|c|c|c|c}
    \hline
     & LINEMOD & LC-HF & Kehl & SSD-6D & \textbf{Ours} \\
    \hline
    Mean &     119 &     n/a  &        671 &         109   & \textbf{43.3}\\
    \hline
   \end{tabular}
 \label{tbl:ACCV3D_time}
 }
 \end{center}
\end{table}

\vspace{3pt} \noindent {\bf Processing time.  }
Mean processing time of the existing methods for ACCV dataset are shown in Table \ref{tbl:ACCV3D_time} (only the time of LC-HF were not provided in the paper).  These timing results are taken from each paper and the algorithms were executed on different computing environments.  However, Kehl and SSD-6D are CNN based methods and were evaluated using GPU (Geforce GTX Titan X for Kehl and GTX 1080 for SSD-6D).  These two methods should have taken more than a few seconds on CPU.  When comparing ours with LINEMOD, ours is faster approximately by three times than LINEMOD on a rather faster CPU (Core i7-2820QM 2.3GHz for LINEMOD and Core i7-7700 3.6GHz for ours) using same number of cores (4 cores).  From these, we can conclude that our proposed method is the fastest among the existing methods.

\begin{figure*}[tb]
 \begin{center}
  \begin{tabular}{c}
   \begin{minipage}{17cm}
    \begin{center}
     \includegraphics[clip, width=3.9cm]{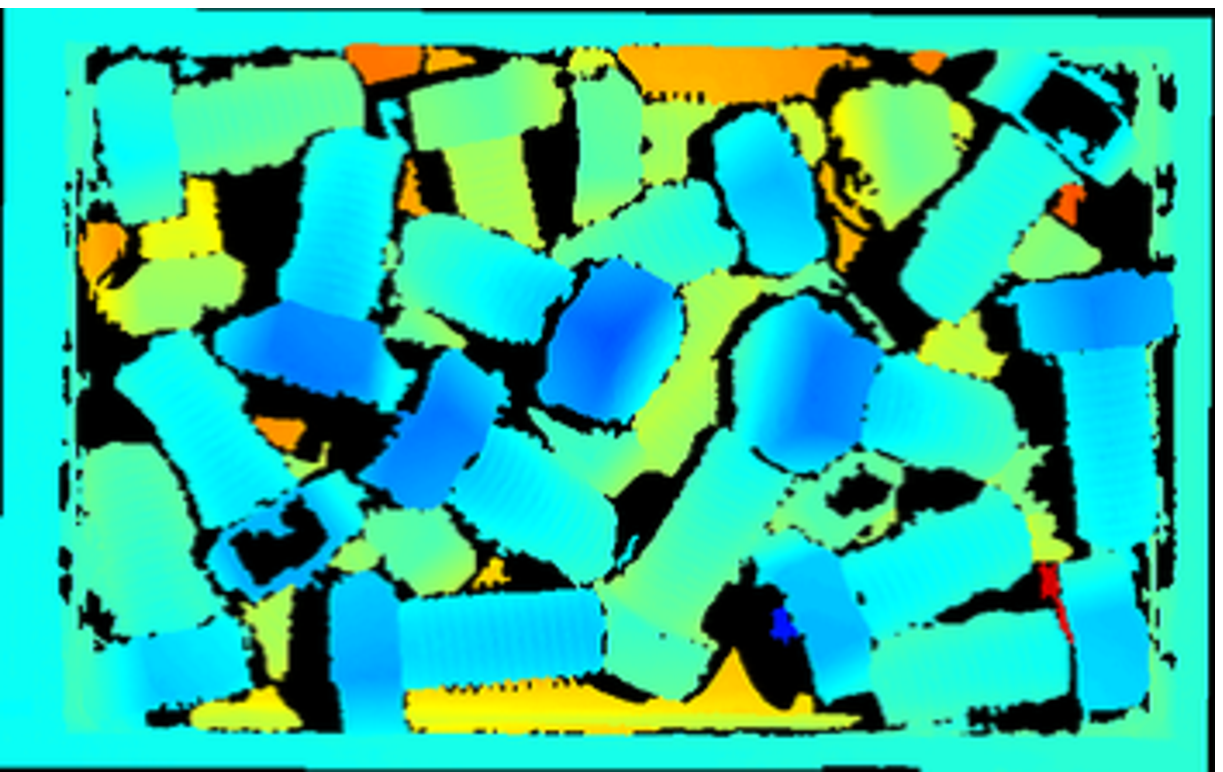}
     \includegraphics[clip, width=3.9cm]{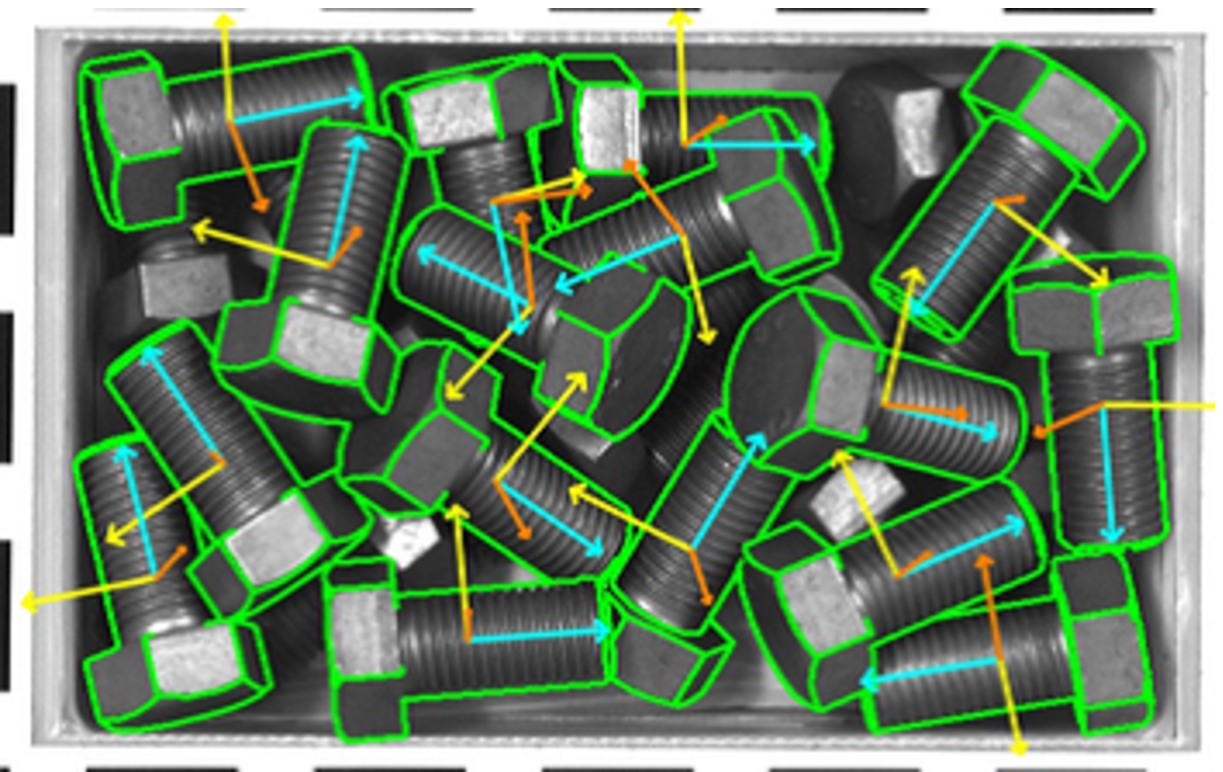}
     \hspace{0.3cm}
     \includegraphics[clip, width=3.9cm]{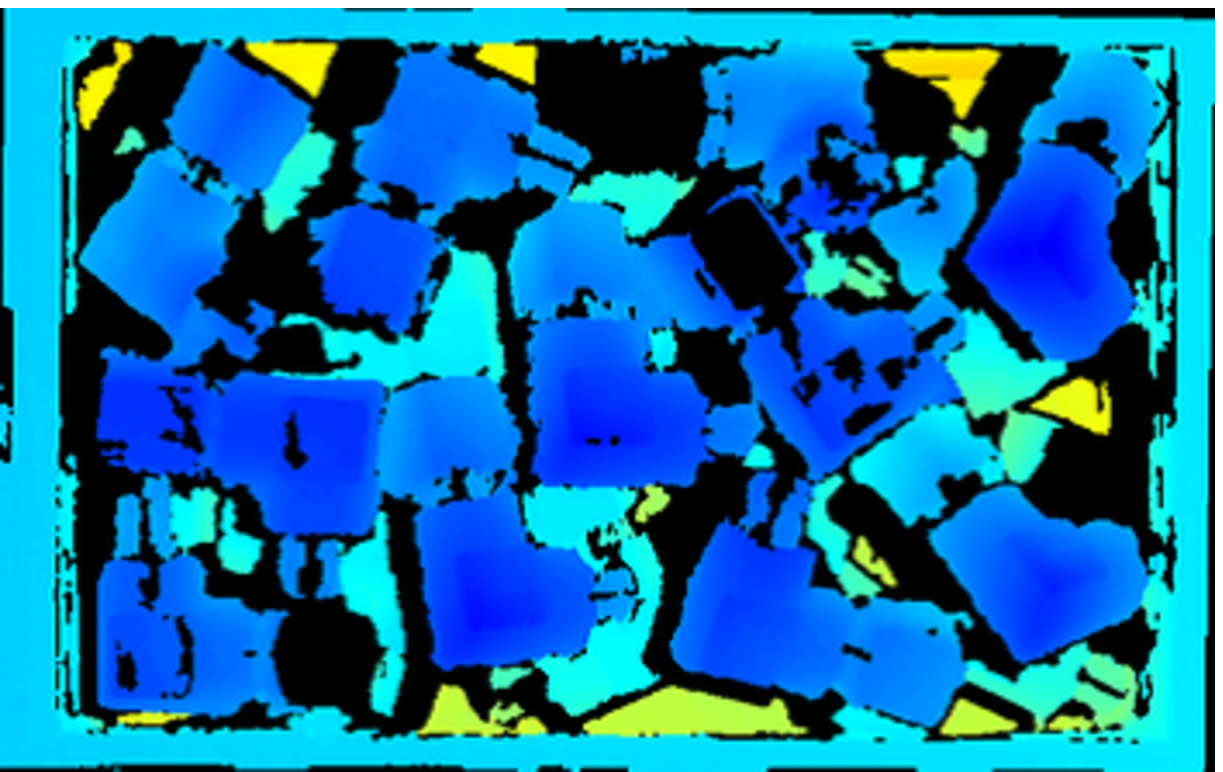}
     \includegraphics[clip, width=3.9cm]{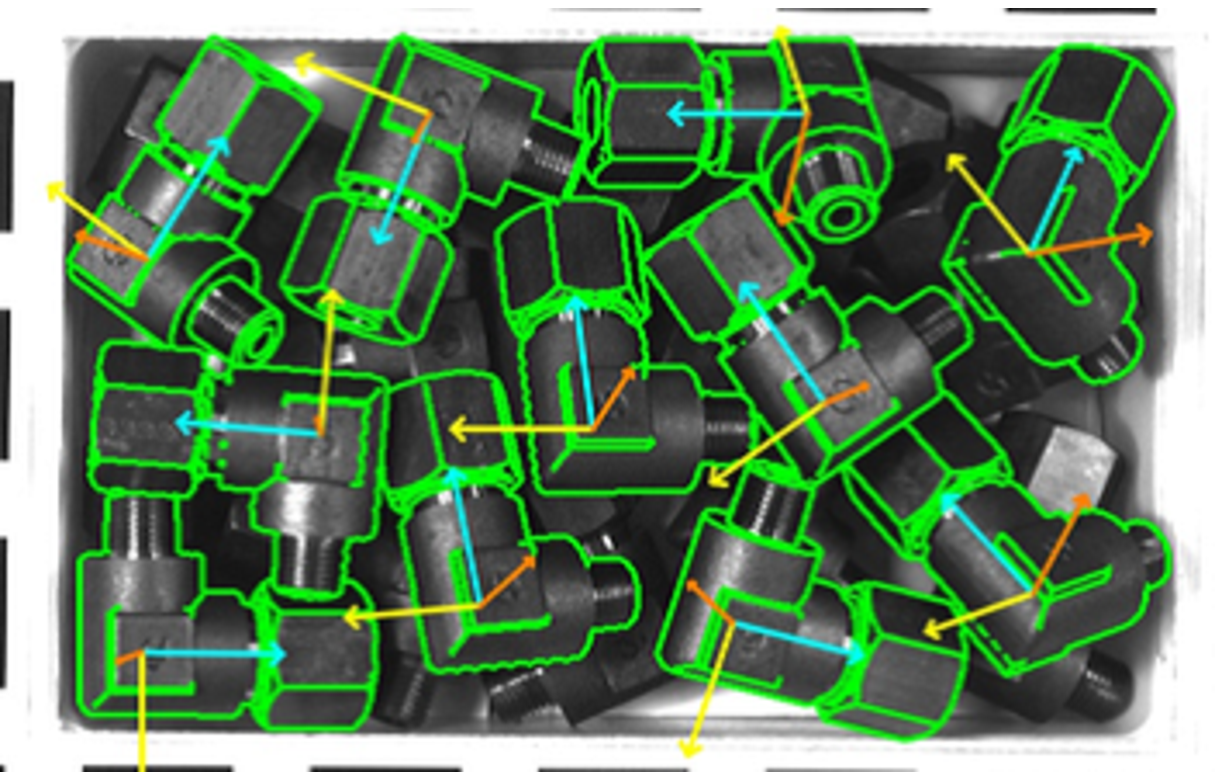}
    \end{center}
   \end{minipage}
  \end{tabular}
  \\
  \vspace{0.1cm}
  \begin{tabular}{c}
   \begin{minipage}{17cm}
    \begin{center}
     \includegraphics[clip, width=3.9cm]{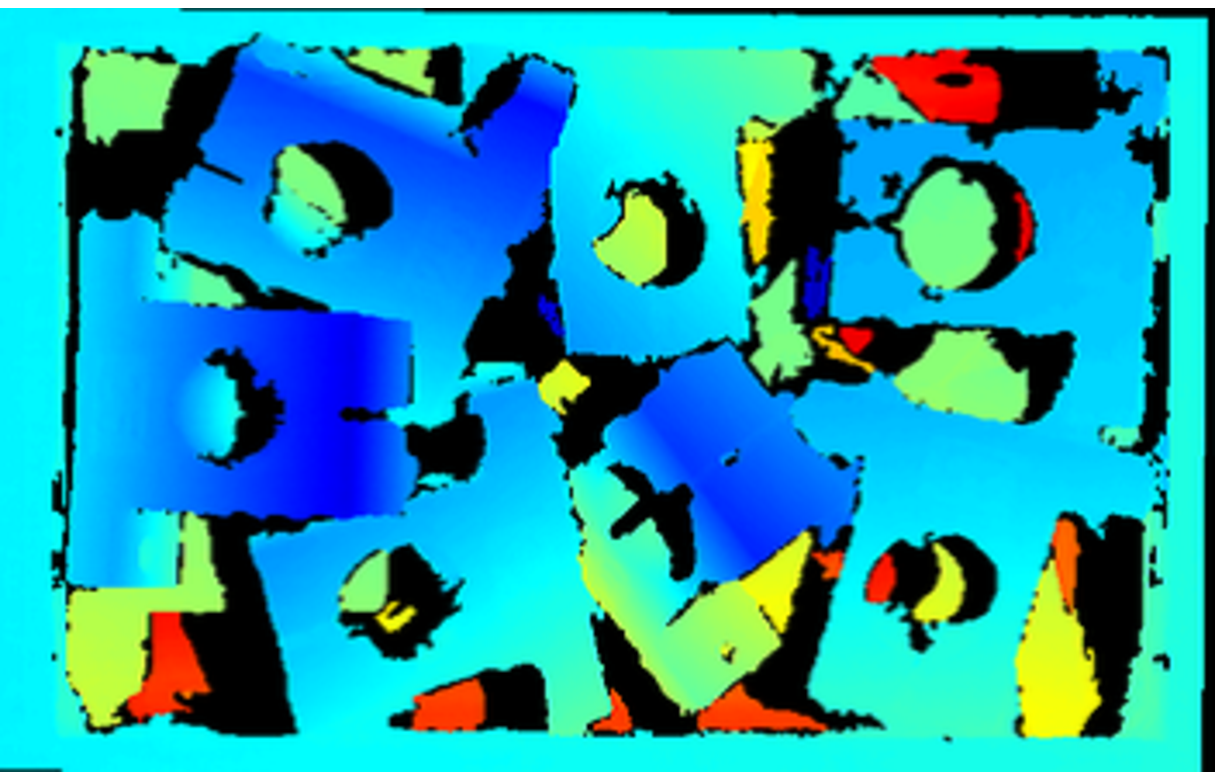}
     \includegraphics[clip, width=3.9cm]{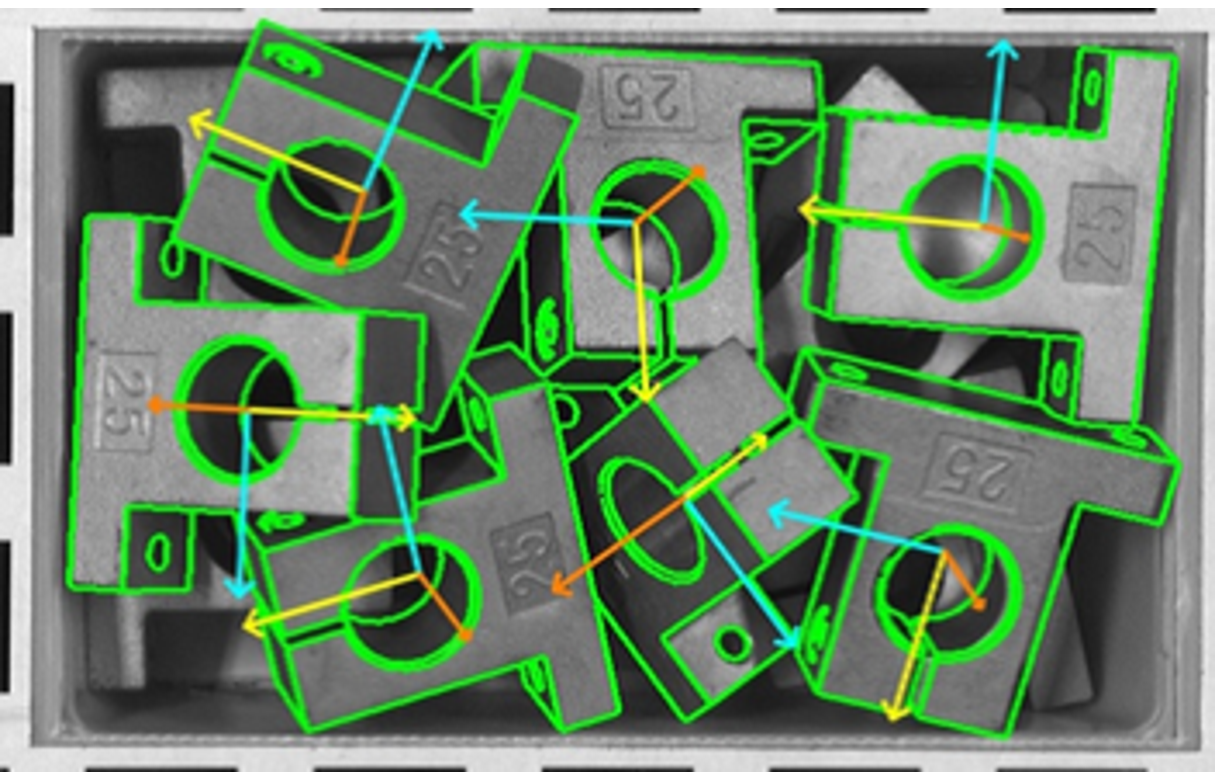}
     \hspace{0.3cm}
     \includegraphics[clip, width=3.9cm]{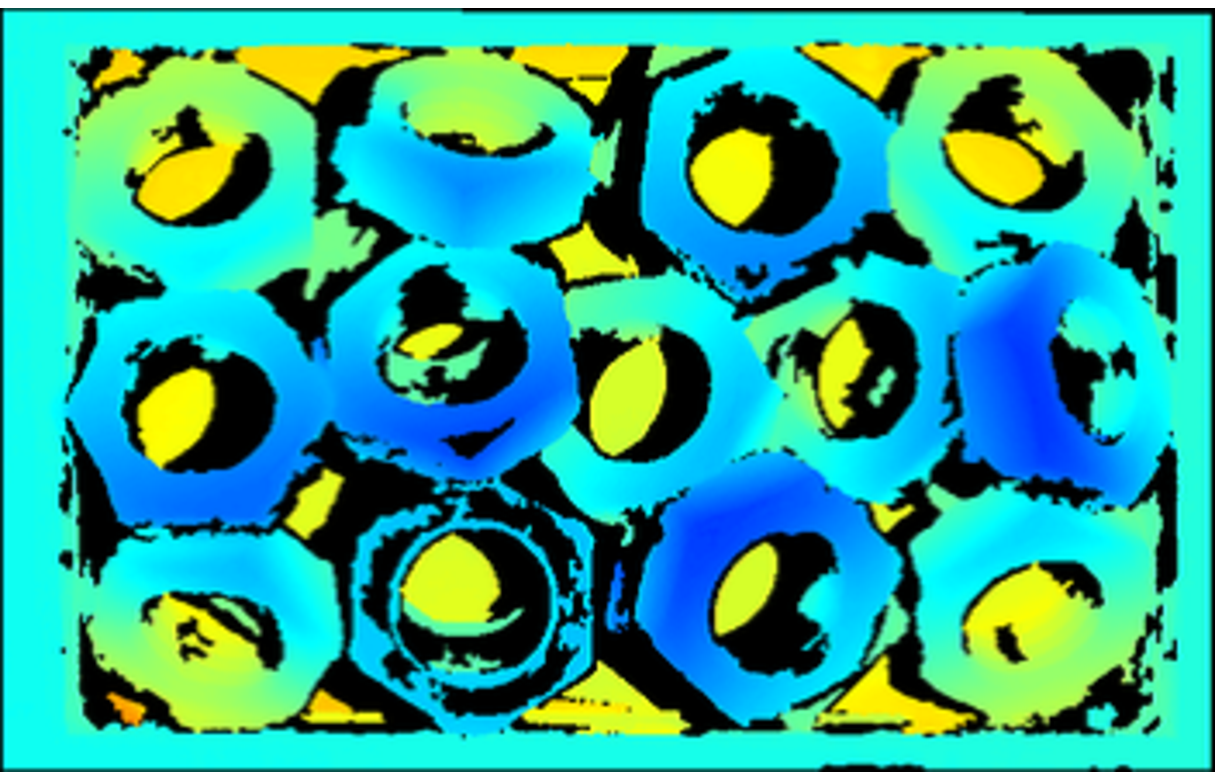}
     \includegraphics[clip, width=3.9cm]{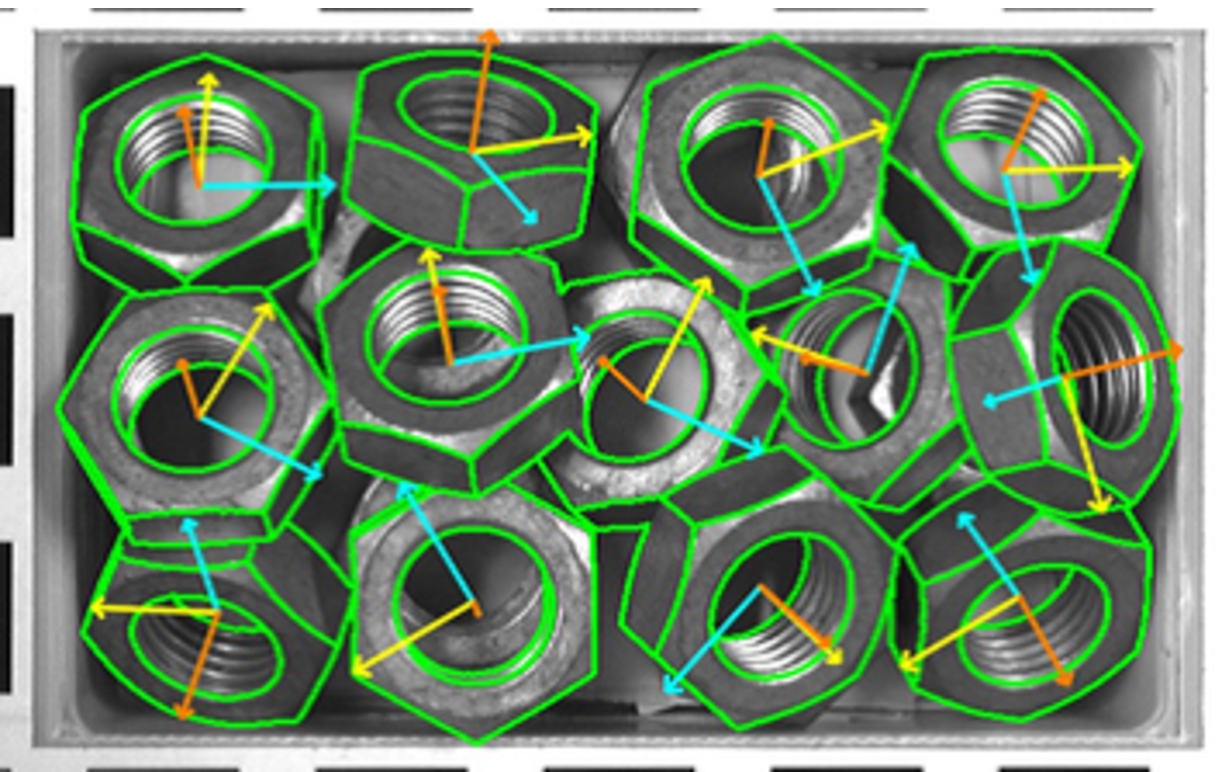}
    \end{center}
   \end{minipage}
  \end{tabular}
  \\
  \vspace{0.1cm}
  \begin{tabular}{c}
   \begin{minipage}{17cm}
    \begin{center}
     \includegraphics[clip, width=3.9cm]{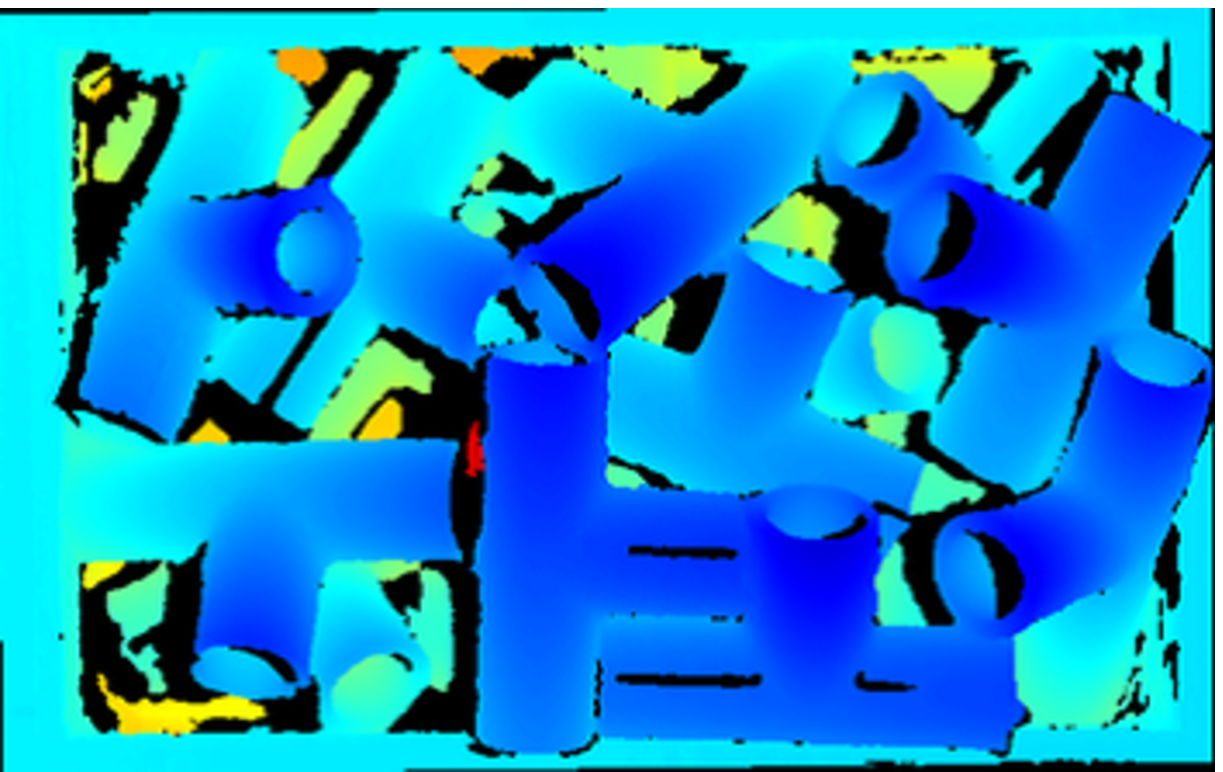}
     \includegraphics[clip, width=3.9cm]{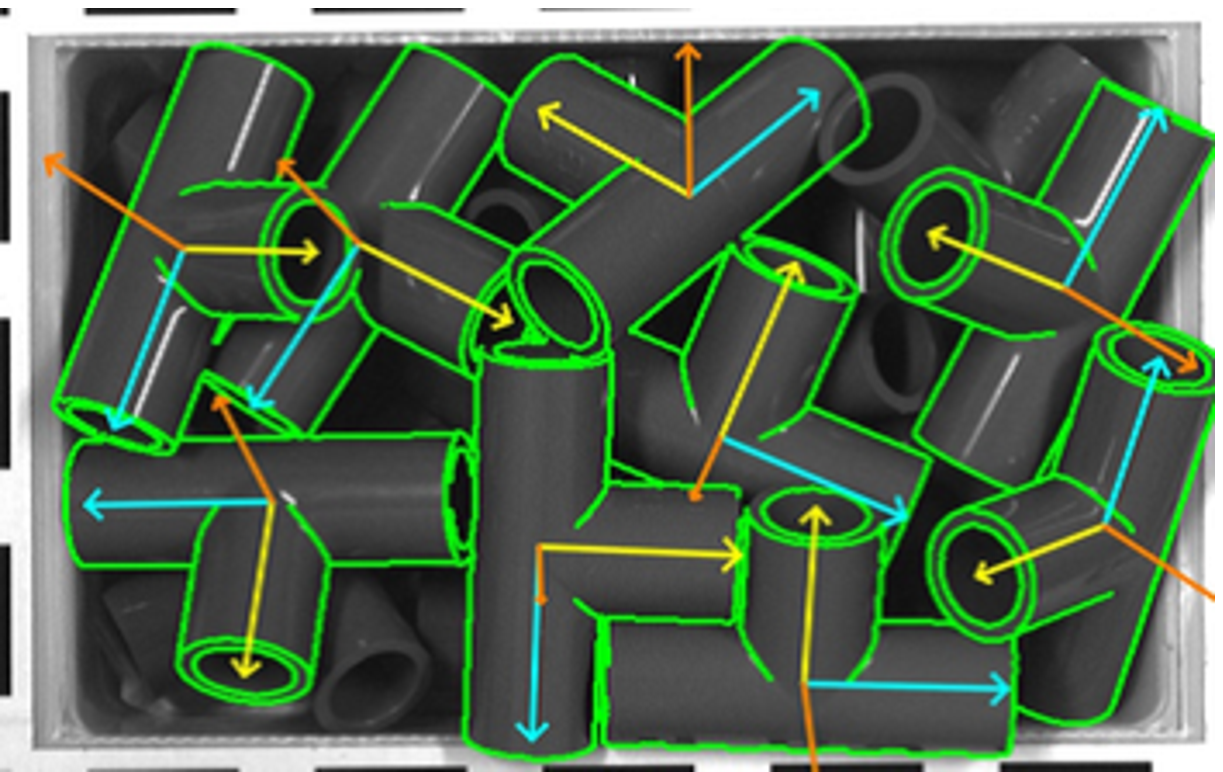}
     \hspace{0.3cm}
     \includegraphics[clip, width=3.9cm]{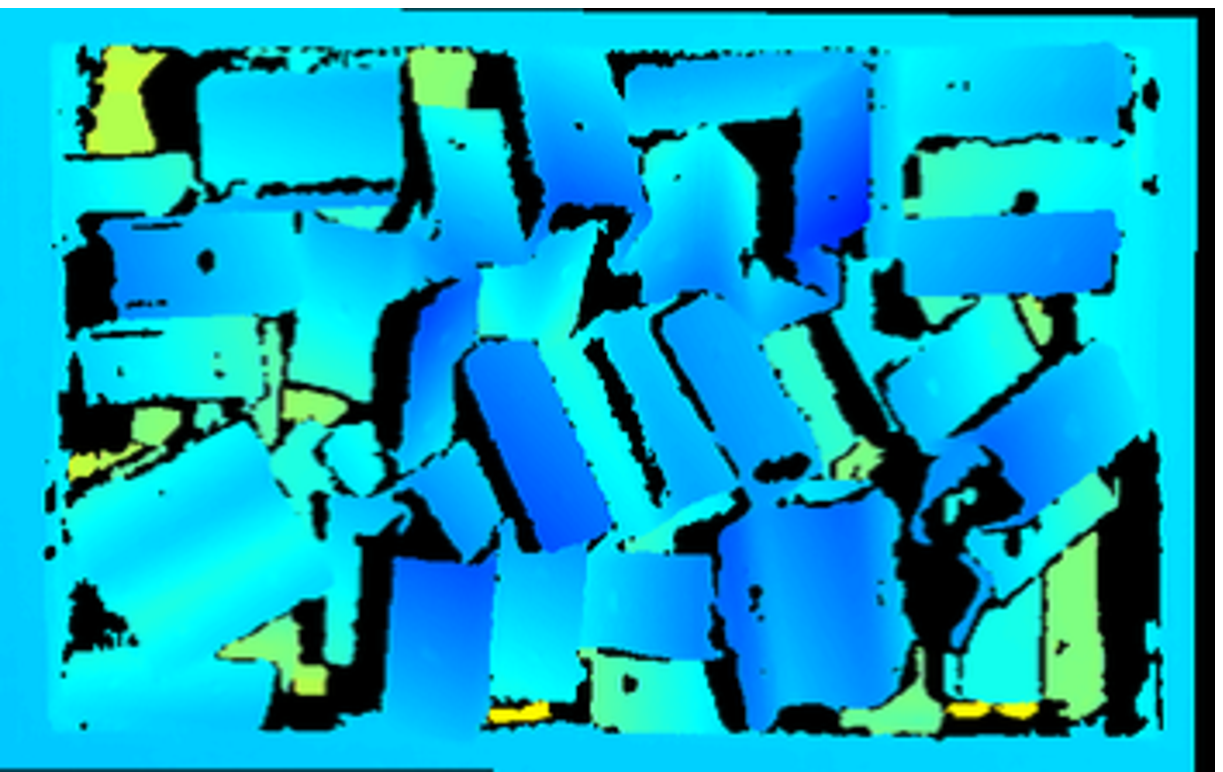}
     \includegraphics[clip, width=3.9cm]{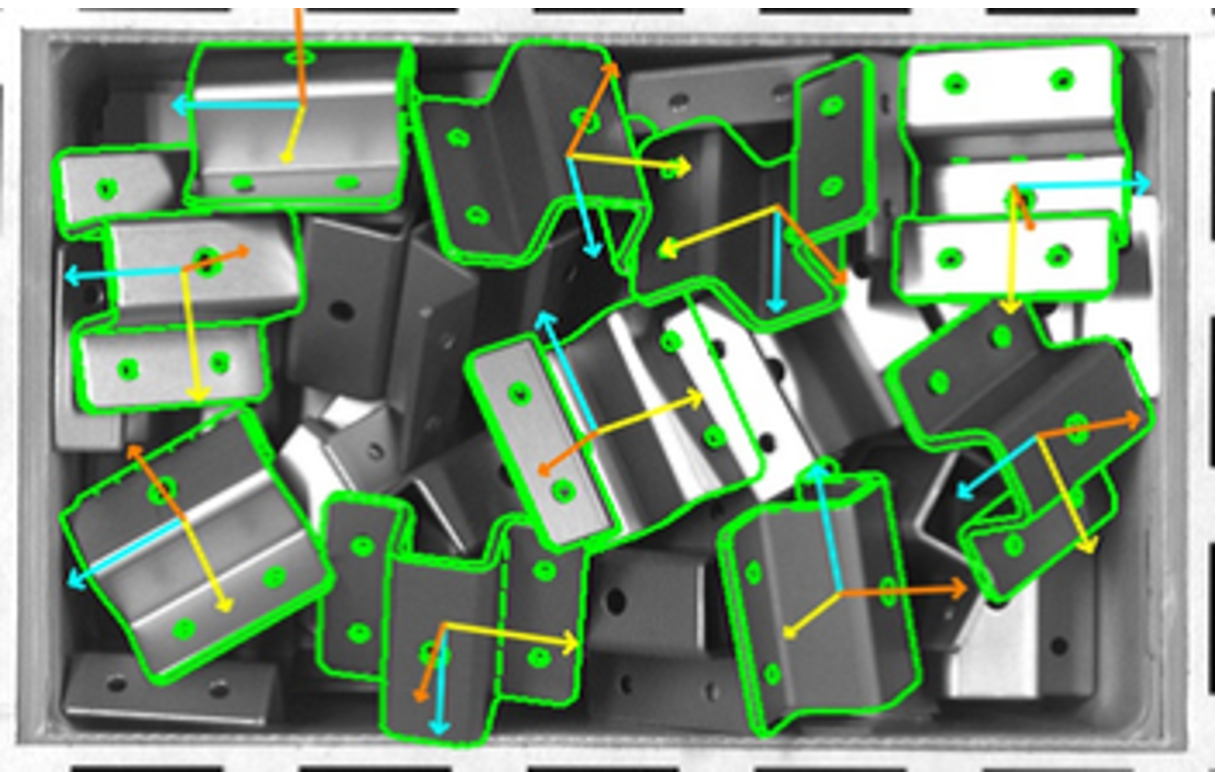}
    \end{center}
   \end{minipage}
  \end{tabular}
 \end{center}
 \caption{Example images of bin-picking dataset.  Top row: Bolt and Connector.  Middle row: Holder and Nut.  Bottom row: Pipe and SheetMetal.  The depth images and grayscale images with our estimation results are shown for each object.}
 \label{fig:ex2_BIN_image}
\end{figure*}

The processing time without the memory rearrangement was 121.7 ms on ACCV dataset.  Therefore, our proposed rearrangement could make 6D pose estimation faster by approximately three times.

%%========================================================%%
%%========================================================%%

\subsection{Evaluation on our bin-picking dataset}
\label{subsec:ex2}
Our algorithm was evaluated also on our bin-picking dataset where the objects were randomly piled in a bin.  The depth and grayscale images ($ 1280 \times 1024$ resolution) of six kinds of mechanical parts were captured using an industrial 3D sensor (Ensenso X36, IDS Gmbh).  A total of 60 images were taken per object which included 5 different patterns of piling, 4 rotation angles of a bin and 3 viewing angles of the 3D sensor.  6D pose of visible target objects (more than 70 \% of the surface are captured by the sensor) were manually annotated and then refined by ICP.  These annotated pose were transformed to the images of the rotated bin and different viewing angles based on AR markers which surrounded the bin.  5 to 10 objects were annotated per bin and the total numbers of annotated objects were 610 (Bolt), 430 (Connector), 210 (Holder), 412 (Nut), 381 (Pipe) and 388 (SheetMetal).  Example images for each object are shown in Figure \ref{fig:ex2_BIN_image}.

Point pair feature (PPF) \cite{Drost2010} was also evaluated on bin-picking dataset.  "Surface-based Matching" which implements PPF on the commercial machine vision software "Halcon13" (MvTec Gmbh in Germany) was used for the evaluation.  Both of PPF and our method used 3D CAD for training and were ran on the same PC (Core i7-7700 3.6GHz) using 4 cores.  The area for pose estimation were limited to inside the bin (approx. $700 \times 400$ pix).

%%--------------------------------------------------------%%
\begin{table}[tb]
 \caption{F1 scores on bin-picking dataset.}
 \begin{center}
  {\tabcolsep = 6mm
   \begin{tabular}{l|c|c}
   \hline
              &  PPF  &  \textbf{Ours}  \\
   \hline
   Bolt       & 0.724 & \textbf{0.888} \\
   Connector  & 0.733 & \textbf{0.921} \\
   Holder     & 0.895 & \textbf{0.950} \\
   Nut        & 0.834 & \textbf{0.922} \\
   Pipe       & 0.850 & \textbf{0.931} \\
   SheetMetal & 0.667 & \textbf{0.774} \\
   \hline
   Mean       & 0.784 & \textbf{0.898} \\
   \hline
   \end{tabular}
  }
 \label{tbl:ex2_BIN_F1}
 \end{center}
\end{table}

\vspace{3pt} \noindent {\bf Estimation accuracy.  }
On our bin-picking dataset, the estimated pose was regarded as correct when the absolute differences of 6 pose parameters (xyz positions and rotation angles) between the estimated pose and the ground truth were smaller than threshold values.  The threshold values of 5 mm in positions and 7.5 degrees in rotations were used.  F1 scores of ours and PPF are shown in Table \ref{tbl:ex2_BIN_F1}.  The scores of ours are higher than those of PPF on all objects.  PPF describes only the shapes of object surfaces and does not explicitly represent the shapes of object contours.  This is why the performance of PPF is degraded when the target objects are industrial parts which commonly consists of simple primitive shapes like planes and cylinders.  The features representing object contours are also helpful for pose estimation of such simple-shaped objects.

%%--------------------------------------------------------%%
\begin{table}[tb]
 \caption{Processing time (ms) on bin-picking dataset.}
 \begin{center}
  {\tabcolsep = 4mm
  \begin{tabular}{l|c|c|c}
  \hline
             &  PPF  & w/o MemRea & w/ MemRea \\
  \hline
  Bolt       & 3334.1 & 239.3 & 117.9 \\
  Connector  & 2382.4 & 236.2 &  99.5 \\
  Holder     & 1343.1 & 104.8 &  47.4 \\
  Nut        & 4387.3 & 307.1 & 139.1 \\
  Pipe       & 1501.5 & 113.5 &  52.1 \\
  SheetMetal & 3121.3 & 196.5 &  86.3 \\
  \hline
  Mean       & 2678.3 & 199.6 &  90.4 \\
  \hline
  \end{tabular}
  }
 \label{tbl:ex2_BIN_time}
 \end{center}
\end{table}

\vspace{3pt} \noindent {\bf Processing time.  }
The processing time of PPF and ours with and without the memory rearrangement are shown in Table \ref{tbl:ex2_BIN_time}.  PPF should be calculated on all pairs of neighboring 3D point clouds.  This leads to longer computation time compared to the per-pixel features like LINEMOD and ours.  Furthermore, our algorithm is accelerated by more than two times when using the memory rearrangement and this brought us 30 times faster speed than PPF.  The time on bin-picking dataset were longer by two times than time on the tabletop datasets because 5 to 10 objects should be detected on bin-picking dataset while 1 to 3 objects on tabletop datasets.

%%--------------------------------------------------------%%
\begin{table}[tb]
 \caption{Mean absolute errors on bin-picking dataset.}
 \begin{center}
  {\tabcolsep = 2.5mm
   \begin{tabular}{l|ccc|ccc}
   \hline
   \multicolumn{1}{l|}{} & \multicolumn{3}{c|}{position (mm)} & \multicolumn{3}{c}{rotation (deg)}\\
   \hline
               &   X   &   Y   &   Z   &   X   &   Y   &   Z   \\
   \hline
    Bolt       & 0.392 & 0.312 & 0.421 & 1.968 & 0.687 & 0.722 \\
    Connector  & 0.443 & 0.391 & 0.448 & 1.372 & 1.302 & 0.991 \\
    Holder     & 0.685 & 0.647 & 0.406 & 0.532 & 0.742 & 1.029 \\
    Nut        & 0.384 & 0.274 & 0.378 & 0.841 & 0.765 & 1.486 \\
    Pipe       & 0.354 & 0.254 & 0.344 & 0.557 & 0.432 & 0.564 \\
    SheetMetal & 0.662 & 0.615 & 0.397 & 0.671 & 0.571 & 0.945 \\
   \hline
    Mean       & 0.487 & 0.415 & 0.399 & 0.990 & 0.750 & 0.956 \\
   \hline
   \end{tabular}
  }
 \label{tbl:ex2_error}
 \end{center}
\end{table}

\vspace{3pt} \noindent {\bf Estimation error.  }
Mean absolute errors of our proposed algorithm for 3D positions (mm) along xyz axes and rotation angles (deg) around xyz axes on bin-picking dataset are shown in Table \ref{tbl:ex2_error}.  These errors are averaged only among the successful results whose errors are less than the threshold values.  The errors in positions are less than 0.5 mm and those in rotations are less than 1.0 degrees.  These are small enough even for robotic manipulation in industrial applications.

%%%%%%%%%%%%%%%%%%%%%%%%%%%%%%%%%%%%%%%%%%%%%%%%%%%%%%%%%%%%
%%%  5.4 Conclusion  %%%%%%%%%%%%%%%%%%%%%%%%%%%%%%%%%%%%%%%
%%%%%%%%%%%%%%%%%%%%%%%%%%%%%%%%%%%%%%%%%%%%%%%%%%%%%%%%%%%%
\section{Conclusion}
\label{sec:Conclusion}
In this paper, we have proposed a real-time 6D object pose estimation from a RGB-D image using only CPU.  Our algorithm is template matching based and consists of three main technical components: PCOF-MOD, BPT and memory rearrangement for a coarse-to-fine search.  The model templates which are made on densely sampled viewpoints and PCOF-MOD which is robust to the changes in object 3D pose contribute to the improved robustness against background clutters.  BPT which is an efficient tree-based data structure and template matching using SIMD instructions on the rearranged feature map make pose estimation faster.  The experimental evaluations on public tabletop and our bin-picking datasets demonstrated that our proposed method was more accurate and faster than the state-of-the-art methods which include recent CNN based methods.

%%%%%%%%%%%%%%%%%%%%%%%%%%%%%%%%%%%%%%%%%%%%%%%%%%%%%%%%%%%%
%%%  References      %%%%%%%%%%%%%%%%%%%%%%%%%%%%%%%%%%%%%%%
%%%%%%%%%%%%%%%%%%%%%%%%%%%%%%%%%%%%%%%%%%%%%%%%%%%%%%%%%%%%


\begin{thebibliography}{99}

\bibitem{Aldoma2011}
A.~Aldoma, M.~Vincze, N.~Blodow, D.~Gossow, S.~Gedikli, R.~Rusu, and G.~Bradski.  Cad-model recognition and 6dof pose estimation using 3D cues.  In {\em Proc. ICCV Workshops}, pages 585--592, 2011.
\bibitem{Birdal2015}
T.~Birdal and S.~Ilic.  Point pair features based object detection and pose estimation revisited.  In {\em Proc. 3DV},  pages 527--535, 2015.
\bibitem{Borgefors1988}
G.~Borgefors.  Hierarchical chamfer matching: a parametric edge matching algorithm.  {\em TPAMI}, 10(6):849--865, 1988.
\bibitem{Brachmann2014}
E.~Brachmann, A.~Krull, F.~Michel, S.~Gumhold, J.~Shotton, and C.~Rother.  Learning 6D object pose estimation using 3D object coordinates.  In {\em Proc. ECCV}, pages 536--551, 2014.
%%\bibitem{Brachmann2016}
%%E.~Brachmann, F.~Michel, A.~Krull, M.~Yang, S.~Gumhold, and C.~Rother.  Uncertainty-driven 6D pose estimation of objects and scenes from a single RGB image.  In {\em Proc. CVPR}, pages 3364--3372, 2016.
\bibitem{Byne1998}
J.~Byne and J.~Anderson.  A CAD-based computer vision system.  {\em Image and Vision Computing}, 16(8):533--539, 1998.
\bibitem{Choi2012b}
C.~C.~Choi and H.~H.~I.~Christensen.  3D textureless object detection and tracking: An edge-based approach.  In {\em Proc. IROS}, pages 3877--3884, 2012.
\bibitem{Cao2016}
Z.~Cao, Y.~Sheikh, and N.~Banerjee.  Real-time scalable 6DoF pose estimation for textureless objects.  In {\em Proc. ICRA}, pages 2441--2448, 2016.
\bibitem{Choi2012a}
C.~Choi, Y.~Taguchi, O.~Tuzel, M.~Liu, and S.~Ramalingam.  Voting-based pose estimation for robotic assembly using a 3D sensor.  In {\em Proc. ICRA}, pages 1724--1731, 2012.
\bibitem{Crivellaro2015}
A.~Crivellaro, M.~Rad, Y.~Verdie, K.~Yi, P.~Fua, and V.~Lepetit.  A novel representation of parts for accurate 3D object detection and tracking in monocular images.  In {\em Proc. ICCV}, pages 4391--4399, 2015.
\bibitem{Cyr2004}
C.~Cyr and B.~Kimia.  A similarity-based aspect-graph approach to 3D object recognition.  {\em IJCV}, 57(1):5--22, 2004.
\bibitem{Damen2012}
D.~Damen, P.~Bunnun, A.~Calway, and W.~Mayol-cuevas.  Real-time learning and detection of 3D texture-less objects: A scalable approach.  In {\em Proc. BMVC}, 2012.
\bibitem{David2005}
P.~David and D.~DeMenthon.  Object recognition in high clutter images using line features.  In {\em Proc. CVPR}, pages 1581--1588, 2005.
\bibitem{Drost2010}
B.~Drost, M.~Ulrich, N.~Navab, and S.~Ilic.  Model globally, match locally: Efficient and robust 3D object recognition.  In {\em Proc. CVPR}, pages 998--1005, 2010.
%%\bibitem{Gavrila2007}
%%D.~Gavrila. A Bayesian, exemplar-based approach to hierarchical shape matching.  {\em TPAMI}, 29(8):1408--1421, 2007.
%%\bibitem{Hinterstoisser2008}
%%S.~Hinterstoisser, S.~Benhimane, V.~Lepetit, P.~Fua, and N.~Navab.  Simultaneous recognition and homography extraction of local patches with a simple linear classifier.  In {\em Proc. BMVC}, 2008.
\bibitem{Hinterstoisser2012a}
S.~Hinterstoisser, C.~Cagniart, S.~Ilic, P.~Sturm, N.~Navab, P.~Fua, and V.~Lepetit.  Gradient response maps for real-time detection of textureless objects.  {\em TPAMI}, 34(5):876--888, 2012.
%%\bibitem{Hinterstoisser2011}
%%S.~Hinterstoisser, S.~Holzer, C.~Cagniart, S.~Ilic, K.~Konolige, N.~Navab, and V.~Lepetit.  Multimodal templates for real-time detection of texture-less objects in heavily cluttered scenes.  In {\em Proc. ICCV}, pages 858--865, 2011.
\bibitem{Hinterstoisser2012b}
S.~Hinterstoisser, V.~Lepetit, S.~Ilic, S.~Holzer, G.~Bradski, K.~Konolige, and N.~Navab.  Model based training, detection and pose estimation of texture-less 3D objects in heavily cluttered scenes.  In {\em Proc. ACCV}, pages 548--562, 2012.
\bibitem{Hinterstoisser2016}
S.~Hinterstoisser, V.~Lepetit, N.~Rajkumar, and K.~Konolige.  Going further with point pair features.  In {\em Proc. ECCV}, pages 834--848, 2016.
%%\bibitem{Hodan2015}
%%T.~Hodan, X.~Zabulis, M.~Lourakis, S.~Obdrzalek, and J.~Matas.  Detection and fine 3D pose estimation of texture-less objects in RGB-D images.  In {\em Proc. IROS}, pages 4421--4428, 2015.
\bibitem{Johnson1999}
A.~Johnson and M.~Hebert.  Using spin images for efficient object recognition in cluttered 3D scenes. {\em TPAMI}, 21(5):433--449, 1999.
\bibitem{Kehl2015}
W.~Kehl, F.~Tombari, N.~Navab, S.~Ilic, and V.~Lepetit.  Hashmod: A hashing method for scalable 3D object detection.  In {\em Proc. BMVC}, 2015.
\bibitem{Kehl2016}
W.~Kehl, F.~Milletari, F.~Tombari, S.~Ilic, and N.~Navab.  Deep learning of local RGB-D patches for 3D object detection and 6D pose estimation.  In {\em Proc. ECCV}, pages 205--220, 2016.
\bibitem{Kehl2017}
W.~Kehl, F.~Manhardt, F.~Tombari, S.~Ilic, and N.~Navab.  SSD-6D: Making RGB-based 3D detection and 6D pose estimation great again.  In {\em Proc. ICCV}, pages 1521--1529, 2017.
\bibitem{Konishi2016}
Y.~Konishi, Y.~Hanzawa, M.~Kawade, and M.~Hashimoto.  Fast 6D pose estimation from a monocular image using hierarchical pose trees.  In {\em Proc. ECCV}, pages 398--413, 2016.
\bibitem{Lanser1995}
S.~Lanser, O.~Munkelt, and C.~Zierl.  Robust video-based object recognition using cad models.  In {\em Intelligent Autonomous Systems IAS-4}, pages 529--536, 1995.
%%\bibitem{Lepetit2009}
%%V.~Lepetit, F.~Moreno-Noguer, and P.~Fua.  EPnP: An accurate o(n) solution to the pnp problem.  {\em IJCV}, 81(2):155--166, 2009.
\bibitem{Rad2017}
M.~Rad and V.~Lepetit.  BB8: A scalable, accurate, robust to partial occlusion method for predicting the 3D poses of challenging objects without using depth.  In {\em Proc. ICCV}, pages 3828--3836, 2017.
\bibitem{RiosCabrera2013}
R.~Rios-Cabrera and T.~Tuytelaars.  Discriminatively trained templates for 3D object detection: A real time scalable approach.  In {\em Proc. ICCV}, pages 2048--2055, 2013.
%%\bibitem{Rusinkiewicz2001}
%%S.~Rusinkiewicz and M.~Levoy.  Efficient variants of the ICP algorithm.  In {\em Proc. 3DIM}, pages 145--152, 2001.
\bibitem{Rusu2010}
R.~Rusu, G.~Bradski, R.~Thibaux, and J.~Hsu.  Fast 3D recognition and pose using the viewpoint feature histogram.  In {\em Proc. IROS}, pages 2155--2162, 2010.
\bibitem{Sundermeyer2018}
M.~Sundermeyer, Z.~Marton, M.~Durner, B.~M., and R.~Triebel.  Implicit 3D orientation learning for 6D object detection from RGB images.  In {\em Proc. ECCV}, 2018.
%%\bibitem{Tanimoto1975}
%%S.~Tanimoto and T.~Pavlidis.  A hierarchical data structure for picture processing.  {\em Computer Graphics and Image Processing}, 4(2):104--119, 1975.
\bibitem{Tejani2014}
A.~Tejani, D.~Tang, R.~Kouskouridas, and T.-K. Kim.  Latent-class Hough forests for 3D object detection and pose estimation.  In {\em Proc. ECCV}, pages 462--477, 2014.
\bibitem{Tekin2018}
B.~Tekin, S.~Sinha, and P.~Fua.  Real-time seamless single shot 6D object pose prediction.  In {\em Proc. CVPR}, 2018.
\bibitem{Tombari2010}
F.~Tombari, S.~Salti, and L.~Di~Stefano.  Unique signatures of histograms for local surface description.  In {\em Proc. ECCV}, pages 356--369, 2010.
\bibitem{Tuzel2014}
O.~Tuzel, M.~Liu, Y.~Taguchi, and A.~Raghunathan.  Learning to rank 3D features.  In {\em Proc. ECCV}, pages 520--535, 2014.
\bibitem{Ulrich2012}
M.~Ulrich, C.~Wiedemann, and C.~Steger.  Combining scale-space and similarity-based aspect graphs for fast 3D object recognition.  {\em TPAMI}, 34(10):1902--1914, 2012.
\bibitem{Wohlhart2015}
P.~Wohlhart and V.~Lepetit.  Learning descriptors for object recognition and 3D pose estimation.  In {\em Proc. CVPR}, pages 3109--3118, 2015.

\end{thebibliography}
\end{document}